\documentclass[journal]{IEEEtran}
\usepackage{amsmath,amsfonts}
\usepackage{algorithmic}
\usepackage{algorithm}
\usepackage{array}
\usepackage[caption=false,font=normalsize,labelfont=sf,textfont=sf]{subfig}
\usepackage{textcomp}
\usepackage{stfloats}
\usepackage{url}
\usepackage{verbatim}
\usepackage{graphicx}
\usepackage{gensymb}
\usepackage[table]{xcolor}
\usepackage{xcolor}
\usepackage{tabularray}
\usepackage{multirow}
\usepackage{array}
\usepackage{soul}
\usepackage[normalem]{ulem}
\usepackage{cite}\makeatletter
\usepackage{placeins}
\usepackage{hyperref}

\begin{document}

\title{\bf MorphoCopter: Design, Modeling, and Control of a \\ New Transformable Quad-Bi Copter}

\author{Harsh Modi, Hao Su,  Xiao Liang, Minghui Zheng
\thanks{This work was partially supported by U.S. National Science Foundation (Grant No.: 2422698).}
\thanks{H. Modi (harsh.modi@tamu.edu) and M. Zheng (mhzheng@tamu.edu) are with the Department of Mechanical Engineering, Texas A$\&$M University, College Station, TX 77843 USA}
\thanks{H. Su (hs5894@nyu.edu) is with the Lab of Biomechatronics and Intelligent Robotics, Department of Biomedical Engineering, Tandon School of Engineering, New York University, New York, NY, 11201, USA.}
\thanks{X. Liang (xliang@tamu.edu) is with the Department of Civil and Environmental Engineering, Texas A$\&$M University, College Station, TX 77843, USA}}
\maketitle

\begin{abstract}
This paper presents a novel morphing quadrotor, named MorphoCopter, covering its design, modeling, control, and experimental tests. It features a unique single rotary joint that enables rapid transformation into an ultra-narrow profile. Although quadrotors have seen widespread adoption in applications such as cinematography, agriculture, and disaster management with increasingly sophisticated control systems, their hardware configurations have remained largely unchanged, limiting their capabilities in certain environments. Our design addresses this by enabling the hardware configuration to change on the fly when required. In standard flight mode, the MorphoCopter adopts an X configuration, functioning as a traditional quadcopter, but can quickly fold into a stacked bicopters arrangement or any configuration in between. Existing morphing designs often sacrifice controllability in compact configurations or rely on complex multi-joint systems. Moreover, our design achieves a greater width reduction than any existing solution. We develop a new inertia and control-action aware adaptive control system that maintains robust performance across all rotary-joint configurations. The prototype can reduce its width from 447 mm to 138 mm (nearly 70\% reduction) in just a few seconds. We validated the MorphoCopter through rigorous simulations and a comprehensive series of flight experiments, including robustness tests, trajectory tracking, and narrow-gap passing tests.

\end{abstract}


\section{Introduction}

\IEEEPARstart{R}{ecent} growth of commercially available drones has enabled widespread adoption in day-to-day life. Applications range from social cinematography to forest-fire detection \cite{forest_fire}, search and rescue \cite{search_rescue}, and medical delivery \cite{zipline}, among others. Among all the drone configurations, the quadcopter in an X arrangement is the most common due to its simplicity and robustness. However, it has a relatively large footprint relative to its payload dimensions, limiting its usability in applications that require these drones to pass through narrow gaps. One such scenario arises in the post-disaster search and rescue. The standard configuration drones must execute precisely planned tilt maneuvers to squeeze through narrow gaps \cite{narrow_gap_1,narrow_gap_2,narrow_gap_3}. During these maneuvers, the drone has to sacrifice lateral and vertical controllability, impeding mission safety and efficiency.

\begin{figure}[t]
      \centering
      \includegraphics[width=0.5\textwidth]{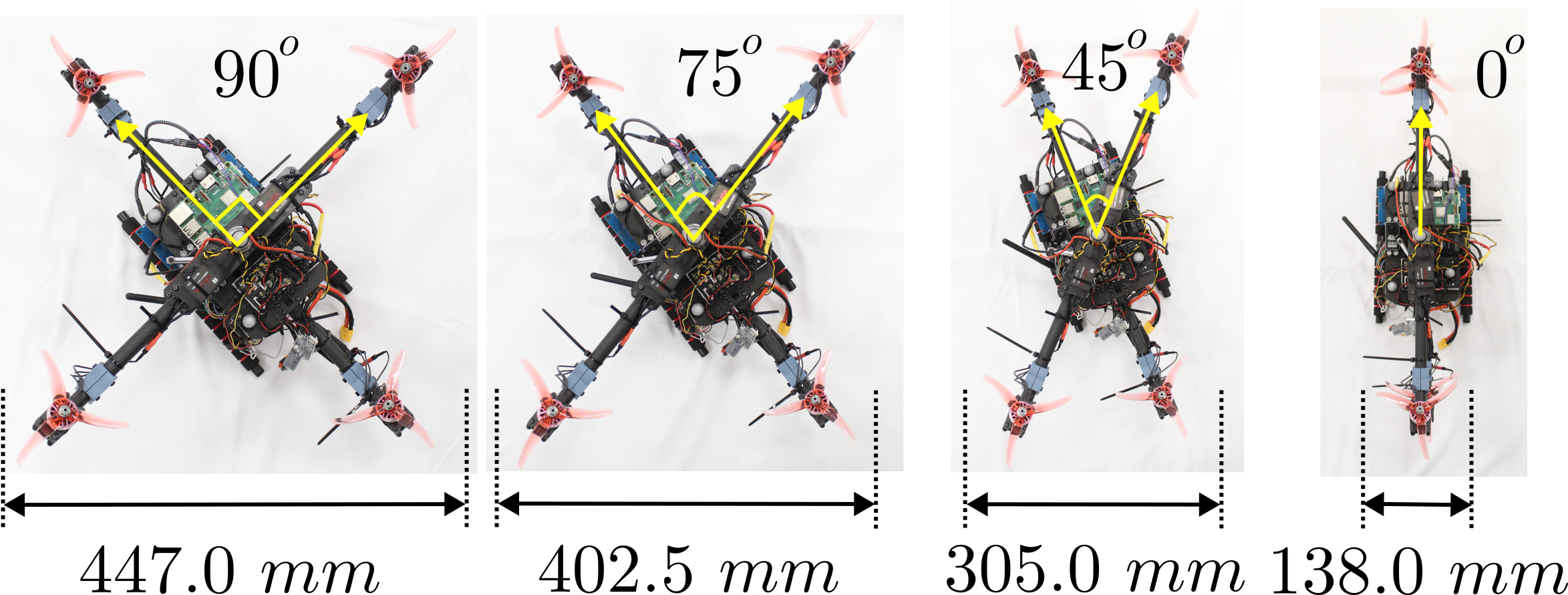}
      \caption{Overview of the MorphoCopter(4-2): where 4 is the number of thrust-producing motors and 2 is the number of rigid links. The figure shows examples of various configurations having different angles between the links and corresponding changes in the width.}
      \label{figure: uav_angles}
   \end{figure}

\IEEEpubidadjcol
To achieve smooth flight through narrow gaps, we require a drone that can morph into a compact configuration temporarily while remaining controllable. It should also be able to revert to the standard configuration for efficient flight.  One such example is the foldable drone that can squeeze and fly through narrow gaps \cite{folding_UAV}. However, this design requires four additional servo motors to control the four arms of the quadrotor independently, making it complex and prone to failures. Although it has the added benefit of controlling the arms independently for complex maneuvers, it does not reduce the width of the drone by a large amount. Similar designs, their dynamic modeling, and their control are available in \cite{folding_UAV2,folding_UAV4,folding_UAV5,folding_UAV6,folding_UAV2_dynamic_modeling,folding_UAV2_gain_scheduling,motion_configuration_control}. Alternatively, the design in \cite{projectile_fold} demonstrates the quadrotor that can rotate the arms containing front two propellers and aft two propellers independently to reduce the width using only two servo motors. However, this design is not controllable in a narrow configuration and merely becomes a projectile while passing through a narrow gap. The drone demonstrated in \cite{airtwin} has two bicopters stacked on top of each other, which can separate in flight and become two independent bicopters. However, this design can only have a standard X-quadcopter configuration or a two-independent bicopters configuration. Also, it cannot go back to the quadcopter configuration after the separation. The design in \cite{soft_drone} uses a compliant mechanism-based quadrotor, which can crawl on the ground through a narrow gap but can not fly through it. Passively morphing quadrotors in \cite{passive_morphing, passive_morphing2} do not require any actuators to morph, but they are not fully controllable during narrow configuration and require precise trajectory planning to provide sufficient momentum to pass through the narrow gap. The shrinking quadrotor demonstrated in \cite{shrinking_UAV} reduces the length of each arm symmetrically using the Sarrus-Linkage mechanism with a maximum size reduction of only $24.4\%$. The flexible arms quadrotor shown in \cite{flexible_UAV} can reduce the size by around $50\%$. However, in that design, the lift force reduces by around $80\%$ in the narrowest configuration, limiting its maneuverability significantly. The design in \cite{tilt_rotor} involves tilting the pairs of rotors such that the roll attitude of the quadrotor body can be controlled independently to very steep angles. However, this design requires 2 additional servo motors, and increases the size in the vertical direction significantly to pass through a narrow gap. There are some other non-conventional designs of multirotor drones in \cite{omni_directional_wrench,multilink_UAV,hoverable_multirotor,3_axis_deformable, ring_rotor,claw_UAV, brick_and_pick}. However, these designs involve complex mechanical and electronic components, which are too expensive and difficult to implement in day-to-day life. The designs in \cite{bicopter,bicopter2} are simple bicopter designs, which are controllable in x-y-z-yaw and can pass through narrow gaps. However, they cannot overcome the simplicity and controllability of the quadcopter configuration in normal flights. 

\begin{table*}[t]
\caption{Comparison among existing designs and our proposed one.} \label{table:comparison}
\begingroup
\setlength{\tabcolsep}{1pt} 
\renewcommand{\arraystretch}{0.0} 

\centering
\begin{tabular}
{m{2.7cm}m{2.2cm}m{2.2cm}m{2.0cm}m{2.2cm}m{2.3cm}m{2.2cm}m{1.8cm}}

\hline

 & 
\centering \includegraphics[width=0.09\textwidth]{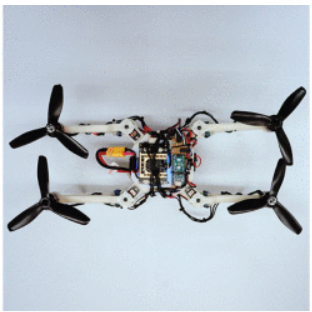} & 
\centering \includegraphics[width=0.09\textwidth]{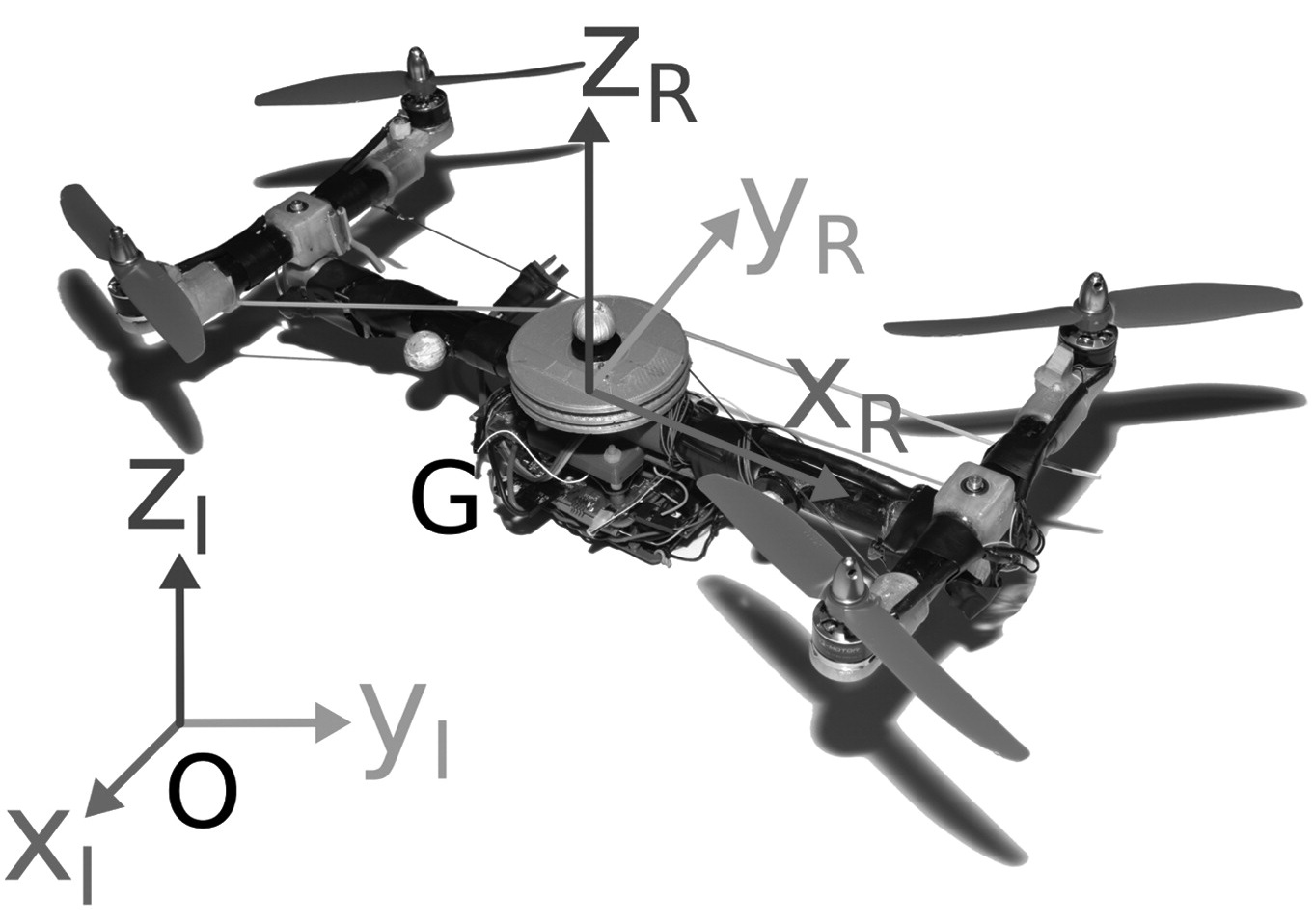} & 
\centering \includegraphics[width=0.09\textwidth]{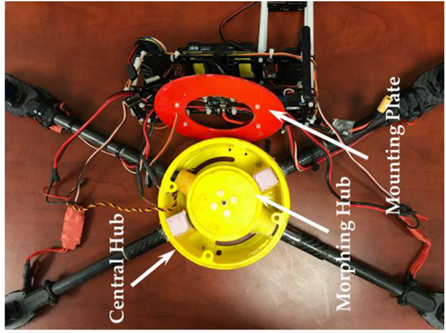} & 
\centering \includegraphics[width=0.09\textwidth]{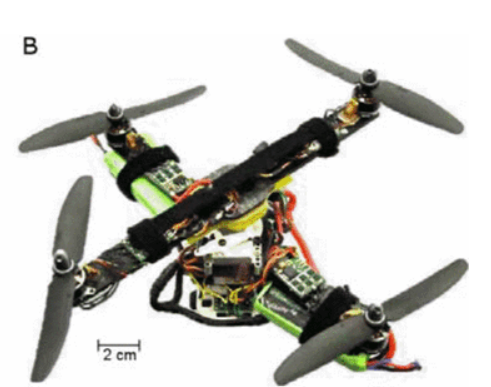} & 
\centering \includegraphics[width=0.09\textwidth]{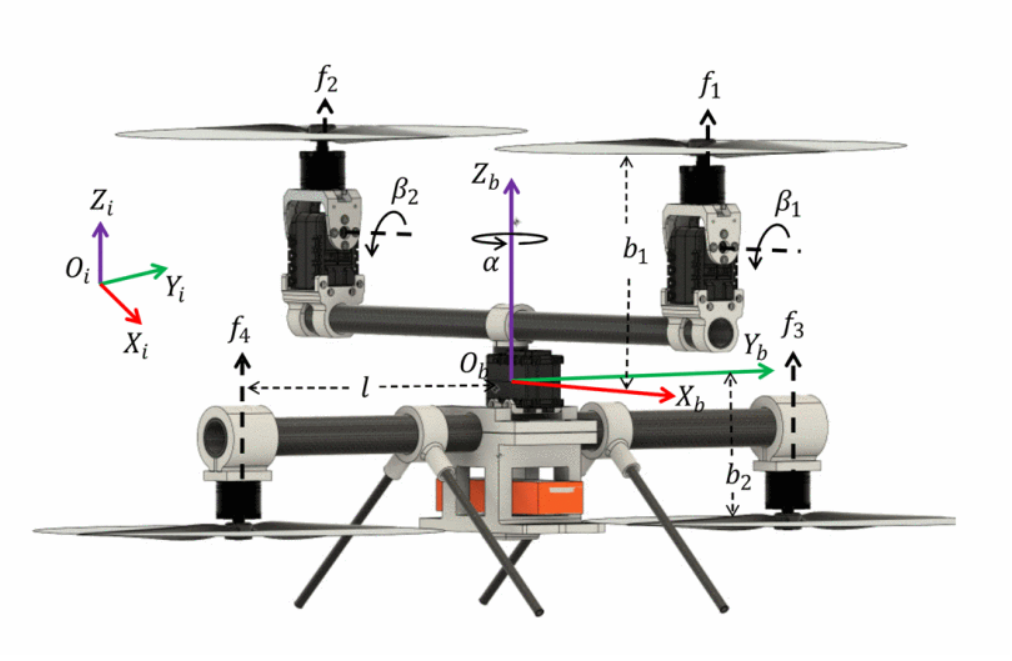} & 
\centering \includegraphics[width=0.09\textwidth]{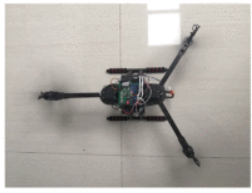} & 
\begin{center}\includegraphics[width=0.09\textwidth]{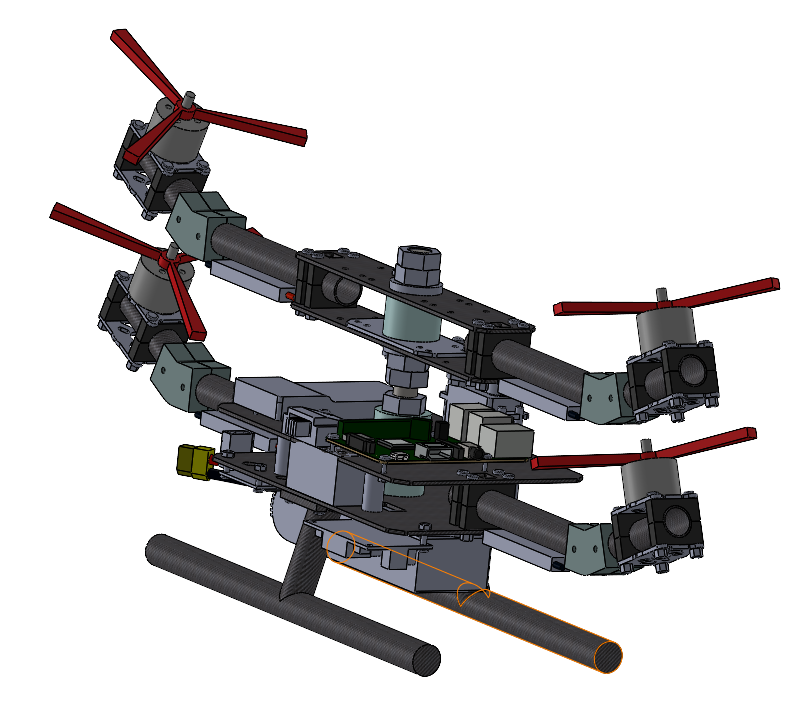} \end{center}\\

\centering \textbf{Design} & 
\centering \textbf{Falanga et al. \cite{folding_UAV}} &
\centering \textbf{Riviere et al. \cite{projectile_fold}} & 
\centering \textbf{Bai et al.\cite{x_morphing_drone}} &
\centering \textbf{Desbiez et al.\cite{x_morphing_drone2}} &
\centering \textbf{Vundela et al.\cite{x_morphing_drone3}} &
\centering \textbf{Hu et al.\cite{y_configuration_passing}} &
\begin{center} \textbf{Ours} \end{center}\\

\rowcolor{gray!20}
\centering Maximum Size Reduction & 
\centering $36.58\%$* & 
\centering $52.24\%$ & 
\centering $15.7\%$* & 
\centering $28.5 \%$ & 
\centering $53.44\%$* & 
\centering $81.13\%$** & 
\begin{center} $\mathbf{69.13\%}$ \end{center} \\

\centering No. of Servos \newline (lower is better)& 
\centering 4 & 
\centering 2 & 
\centering 1 & 
\centering 1 & 
\centering 3 & 
\centering 4 & 
\begin{center} \textbf{1}  \end{center}\\

\rowcolor{gray!20}
\centering Controllable DOFs in Narrow Configuration & 
\centering 4 & 
\centering 0 & 
\centering 4 & 
\centering 4 & 
\centering 4 & 
\centering 4 & 
\begin{center} \textbf{4}  \end{center}\\

\centering Experiments Done & 
\centering Yes & 
\centering Yes & 
\centering Yes & 
\centering Yes & 
\centering No & 
\centering Yes & 
\begin{center} \textbf{Yes} \end{center} \\

\hline
\end{tabular}

\endgroup
~\newline
*: no reported size reduction data available, based on our deduction from figures; **: only half of the drone reduced to this at a time
\end{table*}

To overcome the issues highlighted in the existing designs, we introduce a new morphing drone named MorphoCopter. Our design can be a standard X-quadcopter during a normal flight, which is the most efficient configuration and has almost equal maneuverability in both x and y directions. In addition, to pass through narrow gaps, it can fold in-plane about a central pivot joint, reducing its width continuously until it forms stacked bicopters.
During all these configurations, the morphocopter is fully controllable in x-y-z-yaw. The designs in \cite{x_morphing_drone} and \cite{x_morphing_drone2} are similar to our design but have certain limitations. In \cite{x_morphing_drone}, the quadrotor is normally in an X-configuration with the capability of folding at the center by $21\degree$ and $27\degree$. However, it cannot morph fully into a stacked bicopters configuration, limiting its width reduction capability. Also, it does not delve into controller adaptation due to changes in the configuration. The controller presented in \cite{x_morphing_drone2} uses adaptive control to adjust with the changing dynamics, but their design can also reduce the size of the drone only by $28.5\%$. The design in \cite{x_morphing_drone3} demonstrates a fully folding quadrotor design similar to ours, but it involves the addition of two servo motors just for the attitude control in a stacked bicopters configuration, apart from the servo motor required to modify the configurations. Also, no experimental validation is presented. The design in \cite{y_configuration_passing} can fold into a Y-configuration, making the drone of a propeller width on one side. To pass through extremely narrow gaps, first, the head of the drone is folded into a narrow configuration; then, in the middle of the gap, it has to transform into an inverted Y configuration, making the tail of the drone narrow. This requires very coordinated execution of the trajectory and slows down the process of passing through the narrow gap. The designs that are the closest candidates for passing through a narrow gap are compared in Table \ref{table:comparison}. Compared to these designs, our design reduces the size of the drone by the maximum amount (except design in \cite{y_configuration_passing}, which does not become fully narrow at a time and uses a step-by-step approach to pass), uses the least number of servo motors to control the configurations or attitude, is controllable in narrow configurations and has been proven effective in experiments. To the best of our knowledge, our design achieves the most size reduction while remaining controllable in any configuration without significantly increasing the hardware complexity.

The paper is organized as follows: Section \ref{section:hardware_design} illustrates the mechanical design of the MorphoCopter, Section \ref{section:controller_framework} introduces a moment of inertia and control action-aware controller design for the MorphoCopter. Section \ref{section:simulation and Experiments} shows the extensive simulation and experiment-based performance evaluation. Section \ref{section:conclusion} concludes the paper. Details about the hardware used are shown in Appendix \ref{appendix: hardware_details}.

\section{Hardware Design}
\label{section:hardware_design}

\subsection{MorphoCopter(m-n) Nomenclature}

We introduce a nomenclature in the form of a MorphoCopter(m-n) for future community use, where m indicates the number of thrust-producing motors and n indicates the number of rigid links. For example, we have constructed MorphoCopter(4-2) with four thrust-producing brushless direct current (BLDC) motors with propellers. These motors are mounted on the ends of 2 arms, which can rotate with respect to each other via a revolute joint actuated by a servo motor. These arms also have enough vertical offset to allow the top propellers to overlap with the bottom propellers. For simplicity, we will denote MorphoCopter(4-2) as MorphoCopter hereon. Also, motor and propeller words are used interchangeably to represent the BLDC motor-propeller set. Fig. \ref{figure: uav_angles} showcases a few out of infinitely many possible configurations of the MorphoCopter. These are just some examples, and the MorphoCopter can be controlled at any angle from $0\degree$ to $90\degree$ between arms. In our design, the lower arm also houses a Pixhawk flight controller, an onboard computer Raspberry Pi, a servo motor, and other necessary peripherals. The upper arm only houses the required BLDC motors and electronic speed controllers (ESCs). First, let us define a few geometric notations for MorphoCopter.

   \begin{figure}[thpb]
      \centering
      
      \includegraphics[width=0.4\textwidth]{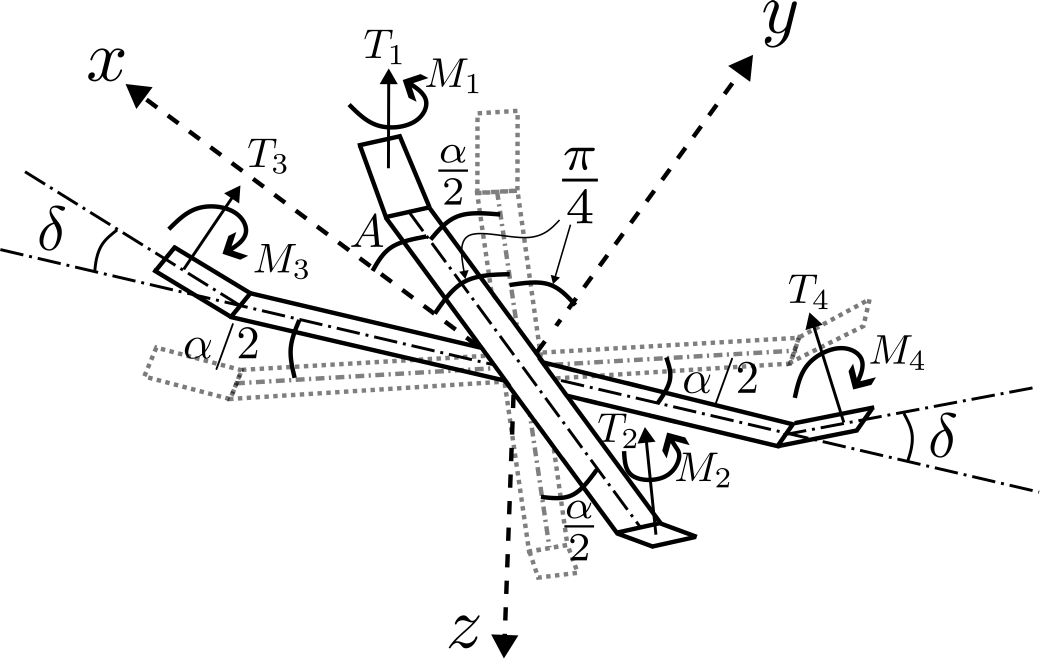}
      \caption{MorphoCopter Geometric Notations. Angles $\alpha$, $A$, and $\pi/4$ are in the x-y plane. Angle $\delta$ is out of the x-y plane} 
      \label{figure: drone_axes}
   \end{figure}

\subsection{MorphoCopter Geometric Notations}

The required notations followed through this paper are shown in Fig. \ref{figure: drone_axes}. We use an FRD (Front - Right - Down) body reference frame. This means the front of the MorphoCopter is the body-attached x-axis, the right of the MorphoCopter is the body-attached y-axis, and downward is the body-attached z-axis. The world reference frame we use is in the NED (North-East-Down) convention. We define the rotation about the body x-axis as roll, about the body y-axis as pitch, and about the body z-axis as yaw. 

\begin{figure}[thpb]
      \centering
      
      \includegraphics[width=0.45\textwidth]{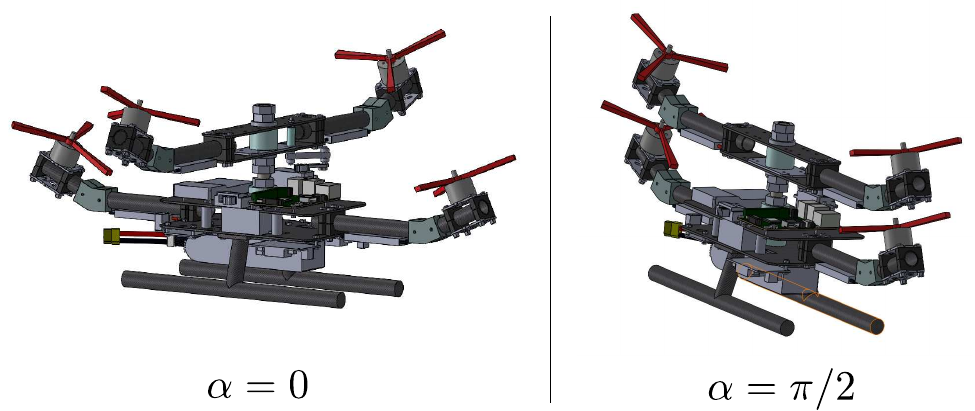}
      \caption{Isometric views of the design with standard quadcopter and stacked bicopters configurations} 
      \label{figure: isometric_views}
   \end{figure}

In Fig. \ref{figure: drone_axes}, $T_i$ indicates the thrust force generated by the motor $i$. $M_i$ indicates the reaction moment generated by the motor $i$. The joint angle $\alpha$ is defined as the rotation of the upper arm with respect to the lower arm, with $\alpha = 0$ being a standard quadcopter configuration and $\alpha = \pi/2$ being a stacked bicopters configuration. Both of these configurations are illustrated in Fig. \ref{figure: isometric_views}. As the upper arm rotates with respect to the lower arm, we consider the body x-axis to always remain at the center of the arms. i.e. when the upper arm rotates with respect to the lower arm, the joint angle $\alpha$ is distributed as $\alpha/2$ on the right side and left side of the MorphoCopter as shown in Fig. \ref{figure: drone_axes}. The dotted lines in Fig. \ref{figure: drone_axes} show where the arms would be in the standard quadcopter configuration, which is $\pi/4$ radians from both the body x and y axes. Hence, the angle made by the upper arm or lower arm with respect to the body x-axis at any joint angle $\alpha$ will be $(\pi/4\ -\alpha/2)$. We will represent this angle as $A$ hereon. The BLDC motors are also tilted inwards by a fixed angle $\delta$ as shown in Fig. \ref{figure: drone_axes} for the roll control in an extremely narrow configuration, which is elaborated in section \ref{subsection: tilt_angle}.

\subsection{Actuated Rotary Joint}
\begin{figure}[thpb]
      \centering
      
      \includegraphics[width=0.45\textwidth]{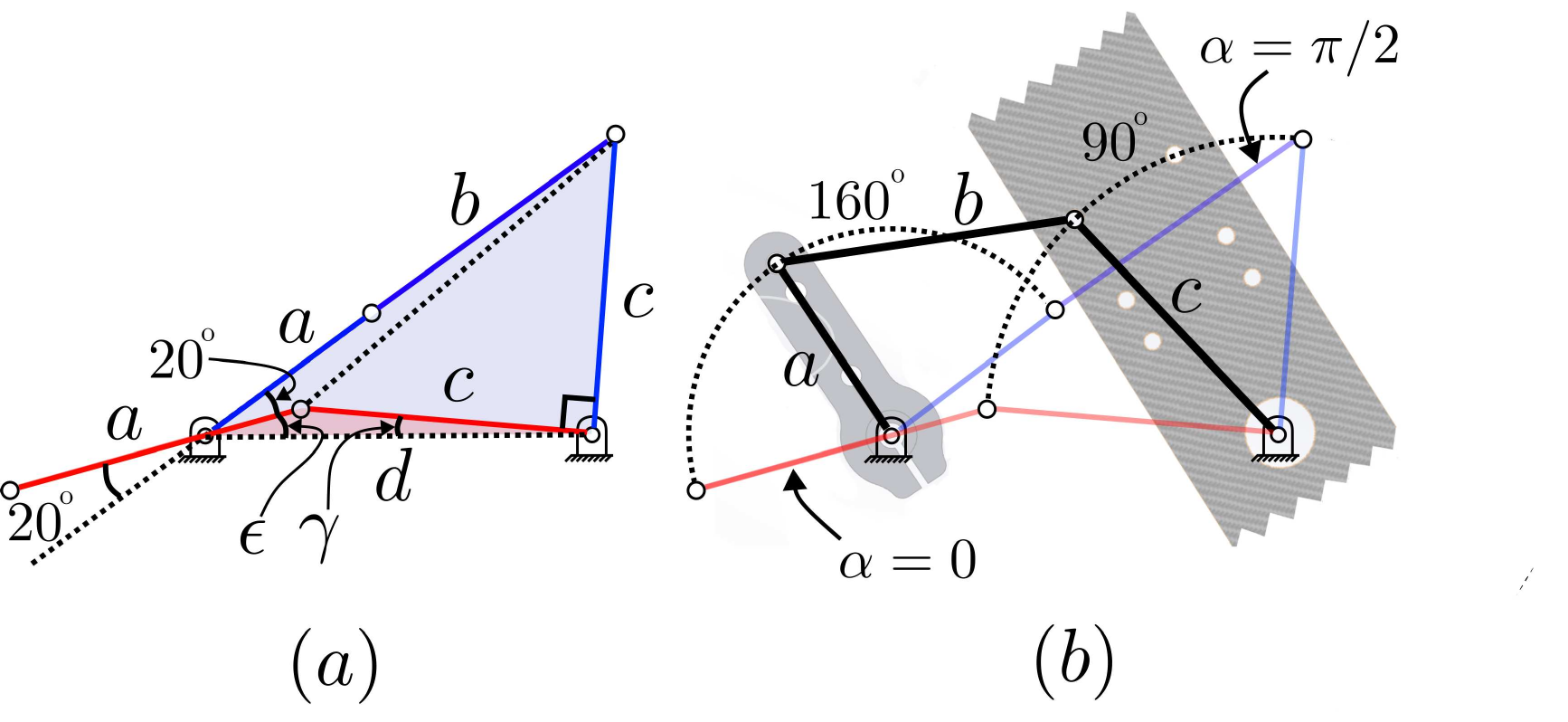}
      \caption{(a) Geometric synthesis of the 4-bar mechanism for upper arm rotation (b) Constructed 4-bar mechanism along with images of servo horn and partial upper arm. Red links correspond to a standard quadcopter configuration ($\alpha =0$) and blue links correspond to a fully folded stacked bicopters configuration ($\alpha = \pi/2)$} 
      \label{figure: 4_bar_combined}
   \end{figure}

A servomotor controls the rotation of the upper arm of the MorphoCopter through a 4-bar linkage mechanism, as shown in Fig. \ref{figure: 4_bar_combined}. The 4-bar linkage parameters are decided considering size constraints, servo motor constraints, and the following requirements to make the hardware robust:
\begin{itemize}
    \item The input angle range should be 160\degree and the output angle range should be 90\degree
    \item The mechanism should ensure that the output is a rocker link.
    \item The mechanism should physically limit the upper arm rotation such that $0 \leq \alpha \leq \pi/2$ is strictly followed even if the servomotor malfunctions.
\end{itemize}

The above requirements ensure that the MorphoCopter does not enter any uncontrollable configuration while maximizing the available leverage of the servomotor torque. We use geometric synthesis to generate a mechanism with the above requirements. Fig. \ref{figure: 4_bar_combined} (a) shows the geometries corresponding to both extremes of the mechanism. Here, $a$ is the servo motor horn, $b$ is a coupler link, $c$ is the distance between the upper arm pivot and the coupler attachment location on the upper arm, and $d$ is the distance between the servo motor axis and the upper arm pivot. The red lines in Fig. \ref{figure: 4_bar_combined} (a) correspond to the joint angle $\alpha = 0$, and blue lines correspond to the joint angle $\alpha = \pi/2$. The angles $\epsilon$ and $\gamma$ are unknown angles that we will use for forming the equations. 

\noindent Now, using the red highlighted triangle shown in Fig. \ref{figure: 4_bar_combined} (a), we can form the following equality:

\begin{small}
\begin{equation}
    \frac{c}{sin\epsilon} = \frac{b-a}{sin\gamma} = \frac{d}{sin(180\degree - \epsilon - \gamma)}
\end{equation}
\end{small}

\noindent Using the blue highlighted triangle in Fig. \ref{figure: 4_bar_combined} (a), we can form the following equality:
\begin{small}
\begin{equation}
    \frac{c}{sin(\epsilon+20\degree)} = \frac{a+b}{sin(\gamma+90\degree)} = \frac{d}{180\degree - ((\epsilon+20\degree)+(\gamma+90\degree))}
\end{equation}  
\end{small}

This system of equations consists of 4 equations and 6 variables. As $a$ is the length of the servo motor horn, we consider it as known (i.e. $24.33\ mm$ in our case). Also, we choose $c$ to be $35\ mm$ as a design choice, considering the feasibility of attaching the coupler to the upper arm. Hence, we solve a system of 4 nonlinear equations with 4 unknowns using the Trust-Region-Dogleg Algorithm. This numerical solution yields the 4-bar mechanism with a = $24.33\ mm$, b = $35.69\ mm$, c = $35\ mm$, and d = $42.48\ mm$. Fig.\ref{figure: 4_bar_combined} (b) shows this mechanism along with extremes of the mechanism corresponding to $\alpha = 0$ and $\alpha = \pi/2$.

\subsection{MorphoCopter Dynamics based Adaptation}
\label{subsection: tilt_angle}
\begin{figure}[thpb]
      \centering
      \includegraphics[width=0.5\textwidth]{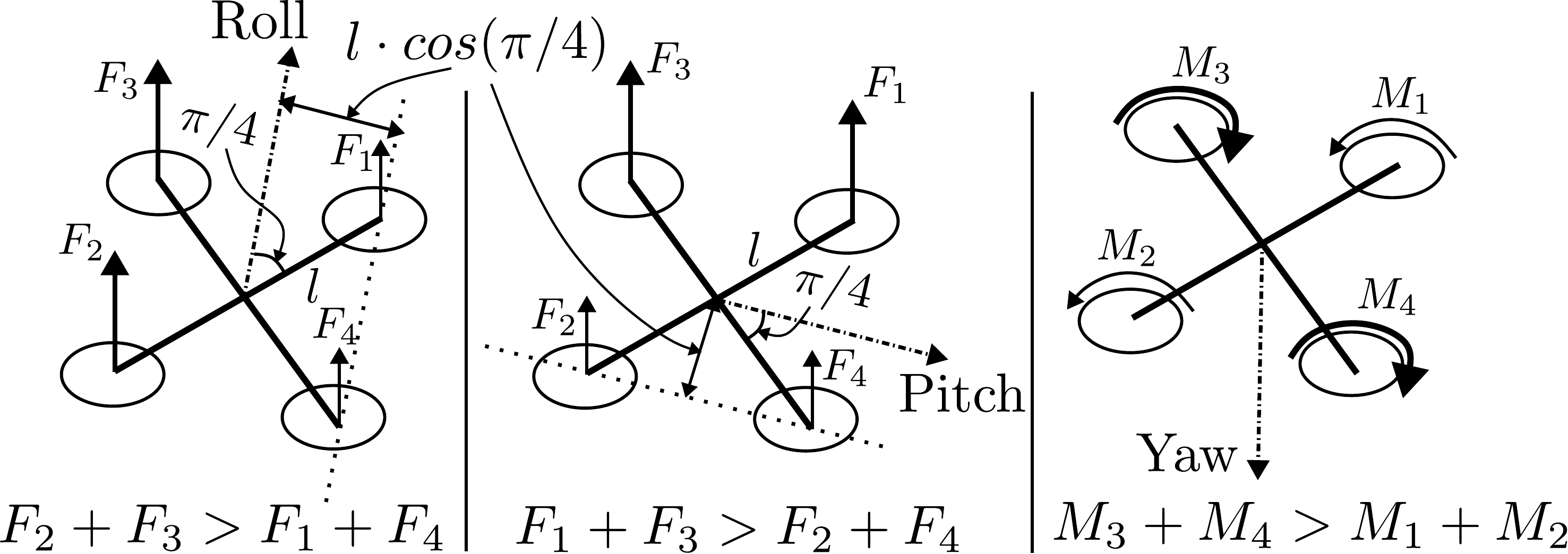}
      \caption{Standard quadrotor attitude control. Differences in thrusts or reaction moments produced are utilized for roll, pitch, and yaw control.} 
      \label{figure: standard_quadrotor_attitude_control}
   \end{figure}

The standard X-configuration quadcopter utilizes the difference in thrusts or reaction moments of propellers to control the attitude as shown in Fig. \ref{figure: standard_quadrotor_attitude_control}. The roll control and pitch control require increasing the total thrust on one side and decreasing it on the other to produce the net moment for the whole body. In the standard configuration, two propellers rotate clockwise, and two propellers rotate counterclockwise. For the yaw control, the speeds of the propellers rotating in the same directions (i.e. clockwise or counterclockwise) are increased or decreased such that the net reaction moment generated contributes towards yaw control. The roll and pitch controls are directly proportional to the distance of the propellers from the center of mass perpendicular to the desired control axis, as indicated by $l\cdot cos(\pi/4)$ in Fig. \ref{figure: standard_quadrotor_attitude_control}. We refer the ability of the propeller thrust manipulation to contribute towards roll/pitch control (proportional to $l\cdot cos(\pi/4)$) as ``control action". In a standard quadcopter, this distance is constant, and hence control-action always remains constant. Also, the rate of change of roll/pitch/yaw angles depends on the respective moments of inertia as well, which remains constant in a standard quadcopter. Hence, a properly tuned controller can remain effective throughout the flight in a standard quadcopter.

\begin{figure}[thpb]
      \centering
      
      \includegraphics[width=0.4\textwidth]{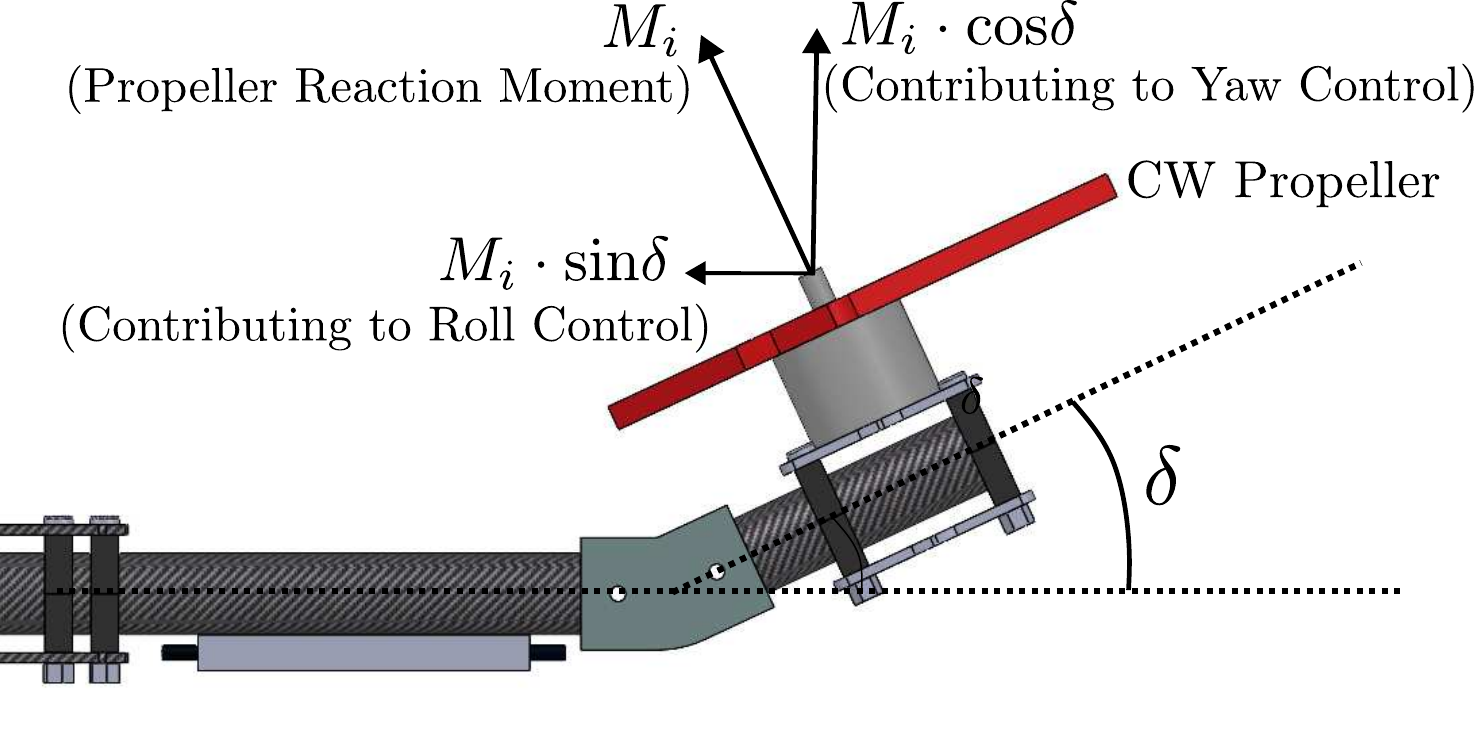}
      \caption{Fixed tilt of motors for roll control for $\alpha > 0$ }
      \label{figure: tilt_side_angle}
   \end{figure}

Now, considering the MorphoCopter design, both the moments of inertia and propeller distances from the center of mass change based on the joint angle. In an extreme folded configuration, in which the MorphoCopter becomes stacked bicopters (i.e. $\alpha=\pi/2$), the distance of propellers from the roll axis becomes 0, and we would completely lose the roll control if the propellers are attached in a standard manner. To address this without increasing the design complexity, we provide a fixed inward tilt ($\delta$) on each motor, as shown in Fig. \ref{figure: tilt_side_angle}. Due to this tilt, the reaction moments generated by the propellers (responsible for yaw control in a standard quadcopter) have a partial component along the roll axis, which can be utilized to control the MorphoCopter. This component along the roll axis increases with increasing tilt angle $\delta$. However, this tilt also reduces the available total upward thrust and yaw control. To determine the required tilt angle $\delta$, we first quantify the moments of inertia and control action of the MorphoCopter. 

\noindent Using Fig. \ref{figure: drone_axes}, the moment of inertia along the body x-axis (i.e. roll axis), with an angle of $A$ between the body x-axis and upper arm (lower arm) can be given by:
\begin{equation}
    \label{equation: Ixalpha}
    \begin{split}
        I_{x}(A) =  (I_{uy}+I_{ly}) sin^2 A+(I_{ux}+I_{lx}) cos^2 A
    \end{split}
\end{equation}

\noindent Where $I_{uy}$ is the moment of inertia of the upper arm along the axis perpendicular to its length and passing through the joint angle pivot. $I_{ux}$ is the moment of inertia of the upper arm along the axis parallel to its length and passing through the joint angle pivot. Similarly, $I_{ly}$ and $I_{lx}$ are the analogous moments of inertia for the lower arm. 

\noindent Similarly, the moment of inertia along the body y-axis (i.e. pitch axis) can be given by:
\begin{equation}
    \label{equation: Iyalpha}
    \begin{split}
        I_{y}(A) =  (I_{uy}+I_{ly}) cos^2 A+(I_{ux}+I_{lx}) sin^2 A
    \end{split}
\end{equation}

\noindent The reaction moments generated by the propellers along with the fixed inward tilt of the propellers (Fig. \ref{figure: tilt_side_angle}) become more responsible for roll control as $\alpha$ increases. The reaction moment generated at the propellers is related to the thrust generated by a moment coefficient $K_m$ defined as:

\begin{equation}
    \label{equation: km}
    M_i = K_m \cdot T_i \approx K_m \cdot \frac{T_{body}}{4}
\end{equation}

\noindent Where $M_i$ is the reaction moment and $T_i$ is the total thrust generated by propeller $i$. $T_{body}$ is the total thrust generated by all the propellers. We will refer $T_{body}$ as $T$ for concise representation. Considering this, the moment generated along the body x-axis (roll axis) by any propeller is given by:

\begin{equation}
    \label{equation: tauxalpha}
        \tau_{x}(A) = \frac{T}{4} cos(\delta) (l\ sinA) + K_m \frac{T}{4} \cdot sin(\delta) cosA
\end{equation}

\noindent Similar moment generated along the pitch axis by a single propeller is given by:

\begin{equation}
    \label{equation: tauyalpha}
        \tau_{y}(A) = \frac{T}{4} cos(\delta) (l\ cosA) + K_m \frac{T}{4} \cdot sin(\delta) sinA
\end{equation}

\noindent Using these quantities, we determine the required inward tilt ($\delta$) of the propellers. If we want to achieve the same body roll rate at joint angles $\alpha = \pi/2$ (i.e. $A = 0$) and $\alpha = 0$ (i.e. $A = \pi/4$), then we need:

\begin{equation}
         \frac{\tau_x(A=0)}{I_x(A=0)} = \frac{\tau_x(A=\pi/4)}{I_x(A=\pi/4)}
\end{equation}

\noindent Using (\ref{equation: tauxalpha}) and (\ref{equation: Ixalpha}), we get:

\begin{small}
\begin{equation}
    \frac{K_m \frac{T}{4} \cdot sin(\delta)}{(I_{ux}+I_{lx})} = \sqrt{2} \cdot \frac{\frac{T}{4} cos(\delta) l + K_m \frac{T}{4} \cdot sin(\delta) }{(I_{uy}+I_{ly}) +(I_{ux}+I_{lx})}
\end{equation}
\end{small}

\noindent Solving this equation for $\delta$ with parameters shown in Table \ref{table:parameters}, we get $\delta = 44.30\degree$. However, at this tilt, the upward component of the total thrust generated from each motor will only be $71.57\%$, reducing the efficiency of the MorphoCopter significantly. So, we decide that we only need half of the maneuverability about the roll axis at configuration with $A = 0$. With this, the required tilt will be calculated using: 

\begin{equation}
         \frac{\tau_x(A=0)}{I_x(A=0)} = \frac{\tau_x(A=\pi/4)}{2 \cdot I_x(A=\pi/4)}
\end{equation}

Solving this equation, we get $\delta = 23.49\degree$. At this tilt, we still have $91.72 \%$ of the total thrust generated in the upward direction, but we get half of the maneuverability compared to the standard quadcopter configuration. For ease of manufacturability, we round off the tilt angle to $25\degree$. Due to the dependency of the moments of inertia and the body moment generation on the joint angle as illustrated in (\ref{equation: Ixalpha}), (\ref{equation: Iyalpha}), (\ref{equation: tauxalpha}), and (\ref{equation: tauyalpha}), it is necessary to design an adaptive controller that considers these factors, which we will discuss next.

\section{Controller Framework}
\label{section:controller_framework} 

\begin{figure}[thpb]
      \centering
      
      \includegraphics[width=0.5\textwidth]{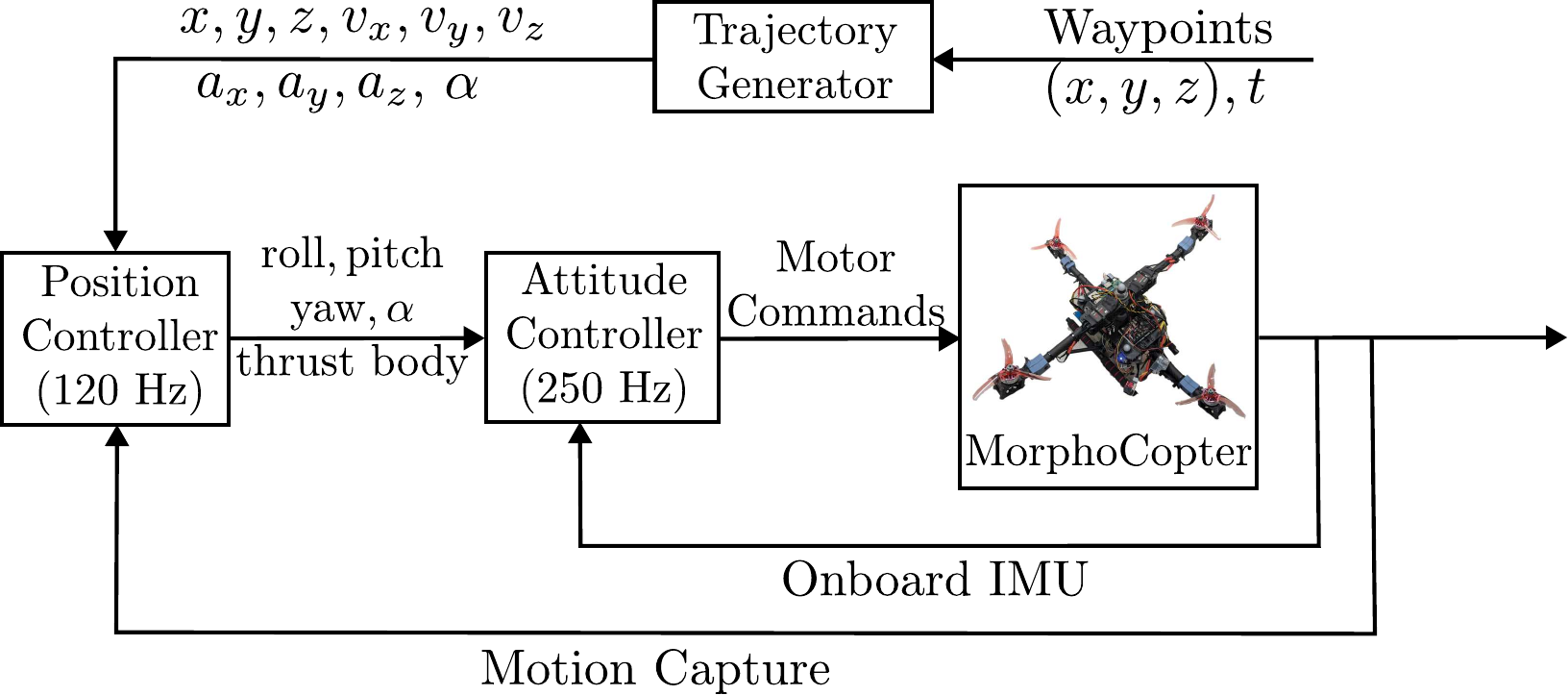}
      \caption{Cascaded Controller} 
      \label{figure: control_loops}
   \end{figure}

The current baseline controller is a nonlinear PID controller implemented in a cascaded format, as shown in Fig. \ref{figure: control_loops}. The waypoints and the desired time to reach those waypoints are provided to the trajectory generator. The trajectory generator generates desired position, velocity, acceleration, and joint angle commands using these waypoints. The trajectory generator currently generates a minimum-snap trajectory of the 5th-order polynomial. Based on these desired values, the position controller calculates the difference between the desired values and the actual values estimated by the motion capture cameras and generates the desired attitude command for the attitude controller. The attitude controller calculates the difference between the desired attitude and the estimated attitude by onboard inertial measurement unit (IMU) sensors and calculates the commands for each of the motors. We will explore these controllers in more detail in upcoming subsections. This baseline controller is an adaptation of the controller formulated in \cite{controller} according to the MorphoCopter varying dynamics.

\subsection{Position Controller}
As shown in Fig. \ref{figure: control_loops}, the position controller takes the desired position, velocity, acceleration, and joint angles as input from the trajectory generator and compares them with motion capture camera-based estimated position and velocity. We calculate error in positions $\mathbf{e_p}$, error in velocity $\mathbf{e_v}$, and integral error $\mathbf{e_i}$. Using these, we can generate the desired force vector $\mathbf{f_{des}}$ using:

\begin{equation}
    \mathbf{f_{des}} = m \cdot \left(\mathbf{a_{des}}+[0,0,-g]^T+\mathbf{k_p}\mathbf{e_p}+\mathbf{k_v}\mathbf{e_v}+\mathbf{k_i}\mathbf{e_i}\right)
\end{equation}

\noindent Here, $m$ is the mass of the MorphoCopter, $\mathbf{a_{des}}$ is the desired acceleration as per the trajectory generator, $g$ is the gravitational constant, and $\mathbf{k_p}$, $\mathbf{k_v}$, and $\mathbf{k_i}$ are position controller gain matrices. Using $\mathbf{f_{des}}$, we can calculate desired body $z$ axis ($\mathbf{z_{b,des}}$) in FRD convention as:

\begin{equation}
    \mathbf{z_{b,des}} = \frac{-\mathbf{f_{des}}}{||\mathbf{f_{des}}||}
\end{equation}

\noindent The intermediate body x-axis ($\mathbf{x_{c,des}}$) based on desired yaw angle is calculated using:

\begin{equation}
    \mathbf{x_{c,des}} = [cos(\psi_{des}),sin(\psi_{des}),0]^T
\end{equation}

\noindent With these, we can determine the desired body y-axis ($\mathbf{y_{b,des}}$) and desired body x-axis ($\mathbf{x_{b,des}}$) as:

\begin{equation}
    \mathbf{y_{b,des}} = \frac{\mathbf{z_{b,des}}\times\mathbf{x_{c,des}}}{||\mathbf{z_{b,des}}\times\mathbf{x_{c,des}}||}
\end{equation}

\begin{equation}
    \mathbf{x_{b,des}} = \mathbf{y_{b,des}}\times\mathbf{z_{b,des}}
\end{equation}

\noindent Using $\mathbf{x_{b,des}}$, $\mathbf{y_{b,des}}$, and $\mathbf{z_{b,des}}$, the desired body rotation matrix with respect to the world frame can be determined using:

\begin{equation}
    \label{equation: Rdes}
    \mathbf{R_{des}} = [\mathbf{x_{b,des}},\mathbf{y_{b,des}},\mathbf{z_{b,des}}]
\end{equation}

\noindent We use this desired rotation matrix and compare it with the estimated rotation matrix in the attitude controller to generate desired motor commands, as explained next.

\subsection{Adaptive Attitude Controller}

Based on the IMU readings, the estimated rotation matrix of the IMU sensor with respect to the world frame is given by:

\begin{equation}
    \mathbf{R_{IMU}} = [\mathbf{x_{b}},\mathbf{y_{b},\mathbf{z_b}}]
\end{equation}

\noindent As shown in Fig. \ref{figure: drone_axes}, as we consider the body x-axis to always remain in the middle of the arms, the body x-axis and IMU sensor (housed in the lower arm) will be at angle $\alpha/2$ with respect to each other. Therefore, the rotation matrix estimated by the IMU sensor needs to be transformed into a MorphoCopter frame of reference using:

\begin{equation}
    \mathbf{R} = \begin{bmatrix} cos(\alpha/2) & sin(\alpha/2)& 0 \\ 
-sin(\alpha/2) &  cos(\alpha/2) &  0\\
0 & 0 & 1\end{bmatrix} \cdot \mathbf{R_{IMU}}
\end{equation}

\noindent Using this estimated rotation matrix and the desired rotation matrix from (\ref{equation: Rdes}), we calculate the error in the attitude of the MorphoCopter using a vee map:

\begin{equation}
    \mathbf{e_R} = \frac{1}{2}\left(\mathbf{R_{des}}\mathbf{R}^T-\mathbf{R}^T \mathbf{R_{des}}\right)^V
\end{equation}

\noindent Using $\mathbf{e_R}$, we can generate integral rotational error using:

\begin{equation}
    \mathbf{e_{R,int}} = \int_{0}^{t} \mathbf{e_R} \,dt
\end{equation}

\noindent To dampen the rotation, we consider the desired angular velocity to be zero, and hence the error in the angular velocity is:

\begin{equation}
    \mathbf{e_W} = -\mathbf{\omega}
\end{equation}

\noindent Where $\mathbf{\omega}$ is the estimated angular velocity by the IMU sensor and then transformed to the MorphoCopter body frame using:

\begin{equation}
    \mathbf{\omega} = \begin{bmatrix} cos(\alpha/2) & -sin(\alpha/2)& 0 \\ 
sin(\alpha/2) &  cos(\alpha/2) &  0\\
0 & 0 & 1\end{bmatrix} \mathbf{\omega_{IMU}}
\end{equation}

\noindent Now, using $\mathbf{e_R}$, $\mathbf{e_{R,int}}$, and $\mathbf{e_W}$, we calculate intermediate desired torque in the body reference frame as:

\begin{equation}
    \label{equation: tau_des_pre_comp}
    \tau'_{des} = \mathbf{k_{p,inner}}\mathbf{e_R}+\mathbf{k_{v,inner}}\mathbf{e_W}+\mathbf{ki_{inner}}\mathbf{e_{R,int}}
\end{equation}

\noindent Where $\mathbf{k_{p,inner}}$, $\mathbf{k_{v,inner}}$, and $\mathbf{k_{i,inner}}$ are the attitude controller gain matrices tuned at $\alpha = 0$. We can determine the desired collective thrust to be generated by all the motors using:

\begin{equation}
    \label{equation: t_body_final}
    T = \mathbf{f_{des}}\cdot \mathbf{z_b}
\end{equation}

\noindent Up to this point, the controller is unaware of the variable moments of inertia and control action of the MorphoCopter. Using the moments of inertia and effective body moments determined in (\ref{equation: Ixalpha}), (\ref{equation: Iyalpha}), (\ref{equation: tauxalpha}), and (\ref{equation: tauyalpha}), the final desired body torque can be calculated from (\ref{equation: tau_des_pre_comp}) as:

\begin{equation}
\label{equation: tau_des_final}
\tau_{des} = \begin{bmatrix} \frac{I_x(A)}{I_{x(0)}}\cdot\frac{\tau_{x(0)}}{\tau_{x}(A)} & 0 & 0 \\ 
0 & \frac{I_y(A)}{I_{y(0)}}\cdot\frac{\tau_{y(0)}}{\tau_{y}(A)} & 0\\
0 & 0 & 1\end{bmatrix}\ \tau'_{des}
\end{equation}

\noindent This final desired torque (Eq. \ref{equation: tau_des_final}) and desired thrust (Eq. \ref{equation: t_body_final}) are sent to the onboard Pixhawk controller running PX4 firmware via its offboard control mode, which then uses its mixing to send the Pulsde-Width-Modulated (PWM) commands to the BLDC motors.

\begin{figure}[thpb]
      \centering
      
      \includegraphics[width=0.4\textwidth]{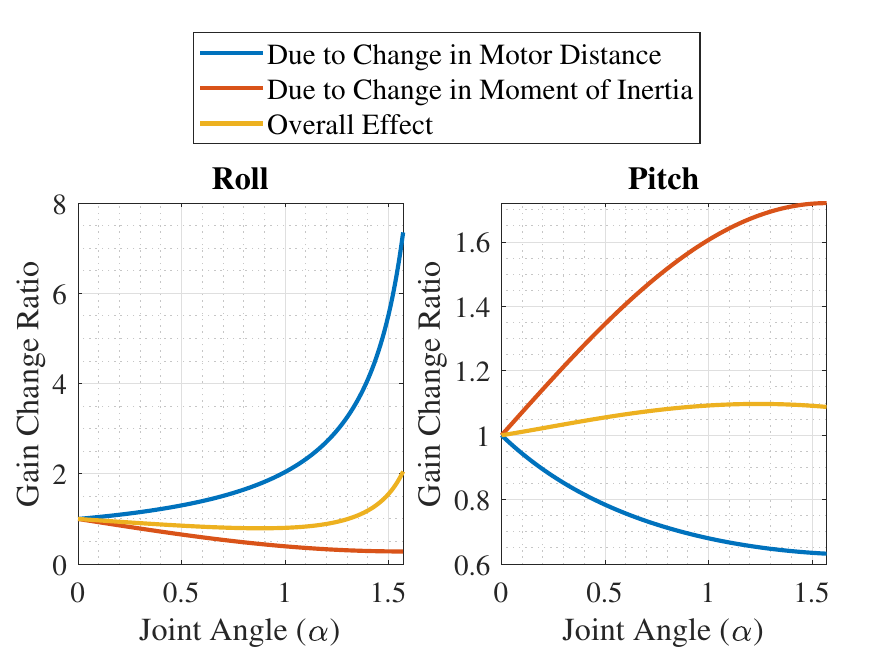}
      \caption{Adaptive gain tuning due to change in motor distance and change in moments of inertia in roll and pitch directions} 
      \label{figure: effort_change_analysis}
   \end{figure}

The ratios in (\ref{equation: tau_des_final}) essentially change the overall gain multiplier of the attitude PID controller. For the prototype of the MorphoCopter we constructed, these ratios affect the attitude controller gains as shown in Fig. \ref{figure: effort_change_analysis}. We can see that in the roll direction, as the joint angle $\alpha$ approaches $\pi/2$, due to propellers and motors coming close to the roll axis, we need to effectively increase the gains by a large amount. However, this decreased motor distance also decreases the moment of inertia about the roll axis. Hence, overall, the gains at $\alpha=\pi/2$ need to be around double the gain at $\alpha = 0$. Also, at some intermediate joint angles ($e.g.\ \alpha = 0.78$), the gains needed in the roll direction are $0.75$ times the gains needed at $\alpha =0$. This makes the MorphoCopter more agile in roll direction at those joint angles. Similarly, in the pitch direction, the moment of inertia increases as $\alpha$ approaches $\pi/2$, but at the same time, control action also increases due to the increased distance between the pitch axis and the propellers. Hence, the overall controller gains remain almost constant in pitch direction for all joint angles.

Next, we will explore the simulations and experiments performed to validate the novel MorphoCopter design and the adaptive controller.

\section{Simulations and Experiments}
\label{section:simulation and Experiments}
The simulation was designed in the Gazebo simulator via PX4 software using the accurate CAD model of the hardware. All the evaluations were first done in the simulator to verify the controller, and then were performed on the hardware in the form of experiments. We primarily performed five kinds of evaluations at various joint angles: A. Hover performance analysis B. External perturbation inputs C. Performance in a shape 8 trajectory D. Performance in a diamond shape trajectory E. Passing through an environment with very narrow gaps. Considering the page limitations and to reduce repetitive details, we show simulation results only for the most complex scenario, while we will show experiment results for all the evaluations. The experiments are also summarized in a video in the supplementary material. 
   
\subsection{Hover performance in experiments}

\begin{figure}[thpb]
      \centering
      
      \includegraphics[width=0.5\textwidth]{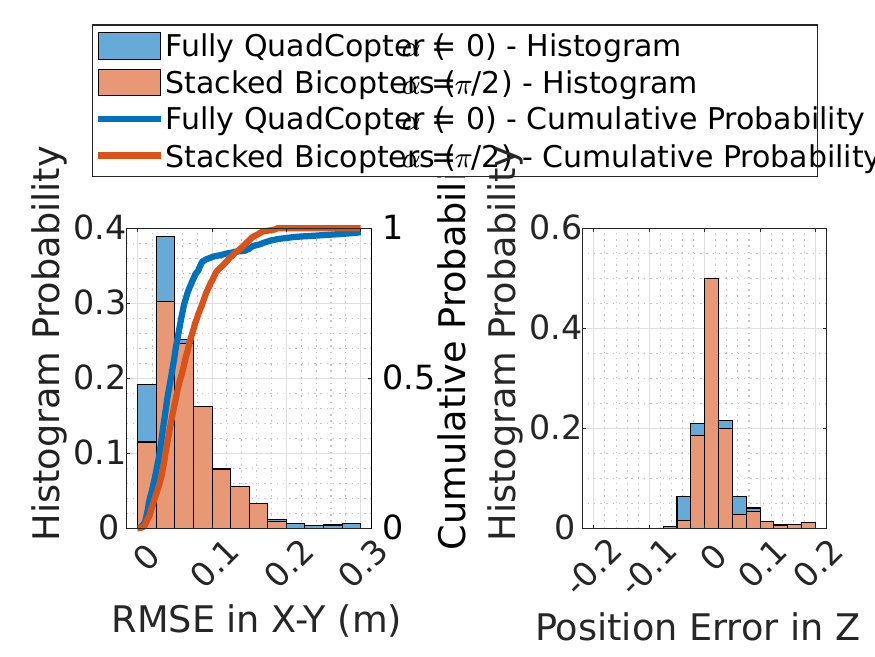}
      \caption{Probability of MorphoCopter hovering within various distances from the desired hover location in experiments} 
      \label{figure: hover_RMSE_histogram}
   \end{figure}

To compare the MorphoCopter's hover performance in standard X configuration ($\alpha =0$) and fully folded stacked bicopters configuration ($\alpha = \pi/2$), experiments with MorphoCopter hovering at the same location ($x = 0\ m$, $y = 0\ m$, $z = -1.2\ m$) were conducted starting with the fully charged battery till the minimum specified battery voltage was reached. The experiments were performed with a 4S 3300 mAh LiPo battery. Without folding, in a standard X configuration, the MorphoCopter could hover for 200.75 seconds. With a fully folded stacked bicopters configuration ($\alpha = \pi/2$), it could hover for 171.56 seconds. Hence, the MorphoCopter could retain almost $85\%$ hover duration in a fully folded configuration compared to the standard quadcopter configuration.

Fig. \ref{figure: hover_RMSE_histogram} shows the hover performance of the MorphoCopter in a histogram format. The figure on the left represents tracking errors in X-Y directions in the form of root mean squared errors (RMSE). This plot also shows the cumulative probability of hovering within various RMSE values using the right axis. The plot on the right shows the histogram of the tracking errors in the Z direction. These plots indicate how well the MorphoCopter can hover close to the desired location. For both configurations, the hover performance is not much different. The cumulative probability plot in the X-Y direction shows that the probability of the MorphoCopter being within 0.1 m RMSE is 0.90 for a standard quadcopter configuration, which is 0.82 for a fully folded stacked bicopters configuration. These plots indicate no significant degradation of hover performance due to the folding of the MorphoCopter.

\subsection{Performance with external perturbations and disturbances in experiments}
\begin{figure}[thpb]
      \centering
      
      \includegraphics[width=0.48\textwidth]{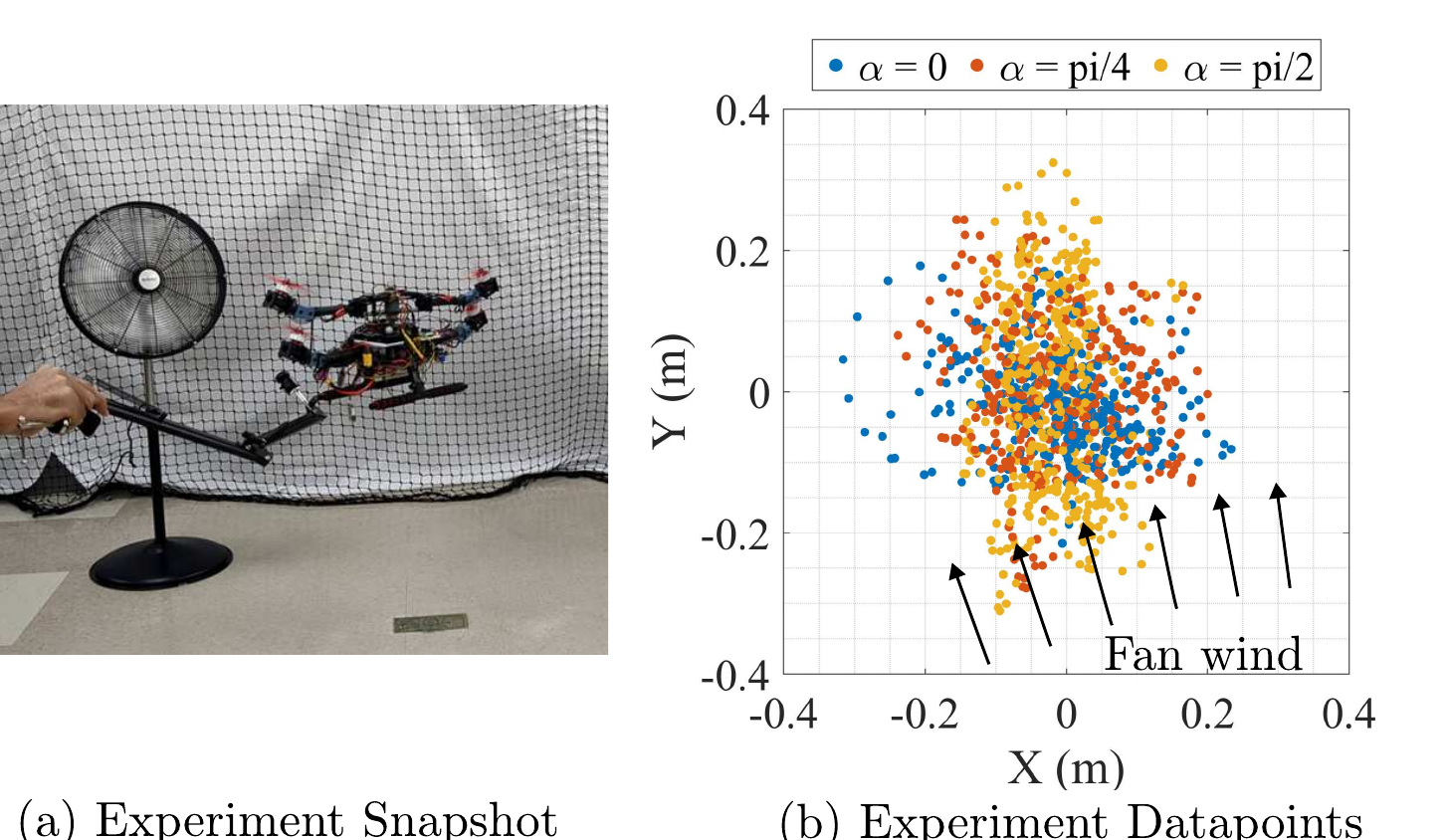}
      \caption{Test of the robustness with manual perturbation and fan winds (a) Snapshot from the experiment video (b) Datapoints of multiple flights at various joint angles} 
      \label{figure: fan_perturbation_XY4c}
   \end{figure}

In this experiment, the MorphoCopter was asked to hover in the same location while subject to changing high winds from a large fan and manually perturbed by a rod. The snapshot from the test video is shown in Fig. \ref{figure: fan_perturbation_XY4c} (a). The fan was kept on the oscillation mode to keep the winds varying with time. Fig. \ref{figure: fan_perturbation_XY4c} (b) shows the data points of the MorphoCopter location in the X-Y plane during the perturbation test, representing the probability density function. The MorphoCopter was able to maintain the hover location mostly within $0.2\ m$ distance from the target hover location in any configuration. This evaluation can be better visualized in the supplementary video.

\subsection{Performance in a shape 8 trajectory in experiments}
\begin{figure}[thpb]
      \centering
      
      \includegraphics[width=0.5\textwidth]{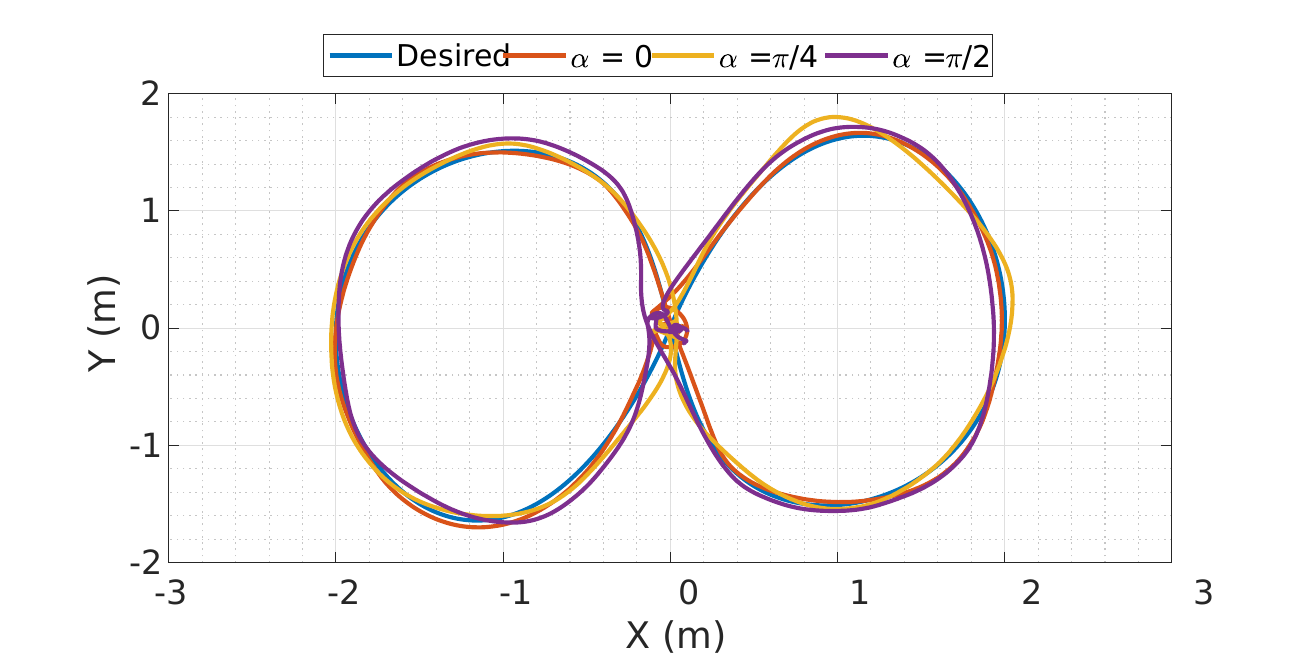}
      \caption{Shape 8 trajectory at various joint angles in experiments} 
      \label{figure: shape8_xy_plot}
   \end{figure}

To verify the controllability of the MorphoCopter in complex maneuvers, the experiments with the shape 8 trajectory were performed at various joint angles. This trajectory was a constant altitude (i.e. $-1.2\ m$) with X-Y trajectory in shape 8 as shown in Fig. \ref{figure: shape8_xy_plot}. We performed the experiments with joint angles $\alpha = 0$, $\alpha = \pi/4$ and $\alpha = \pi/2$. As shown in Fig. \ref{figure: shape8_xy_plot}, the MorphoCopter is able to follow this complex trajectory well in all the configurations. 

\begin{figure}[thpb]
      \centering
      
      \includegraphics[width=0.38\textwidth]{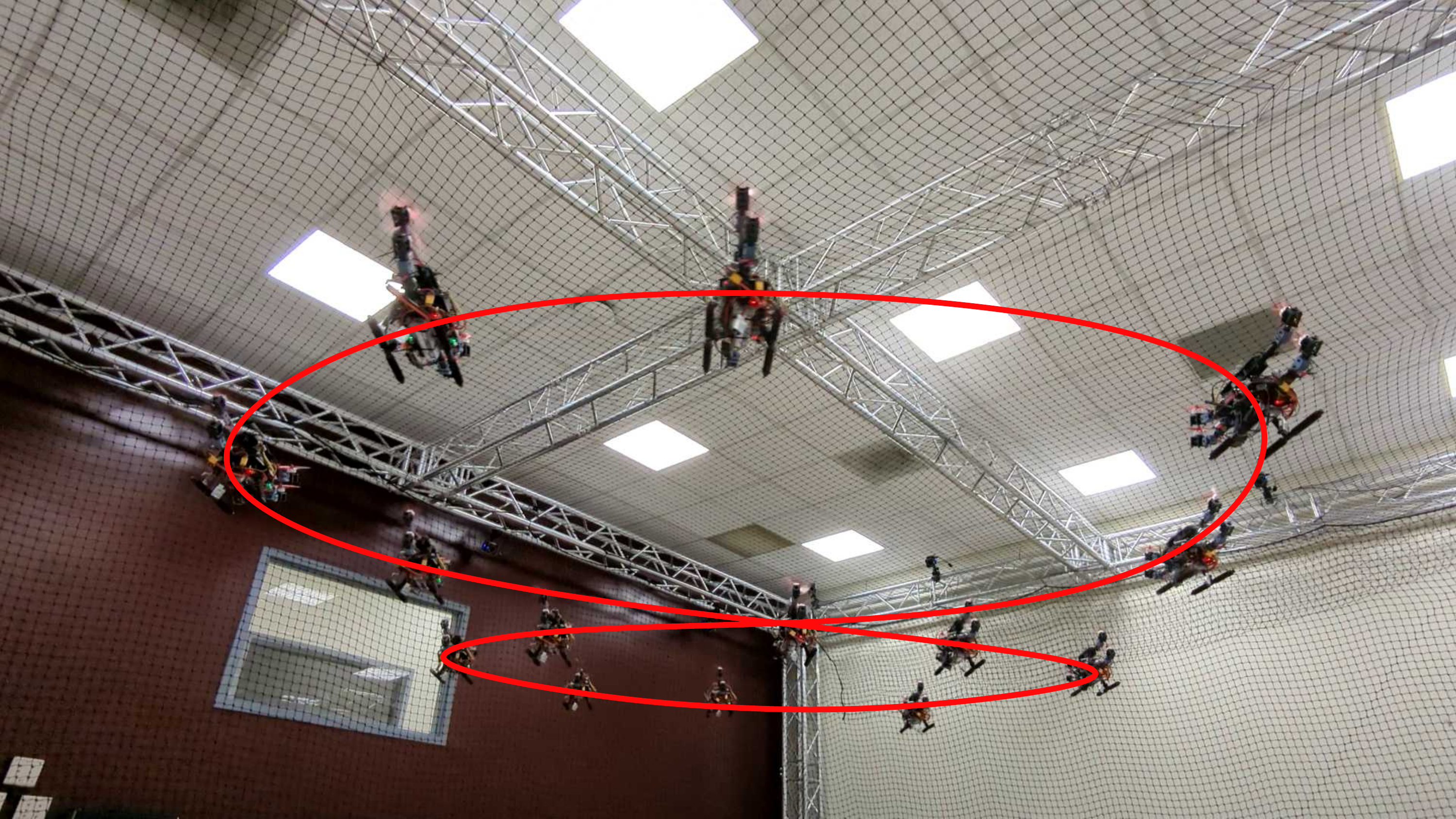}
      \caption{Snapshots from shape 8 trajectory traverse with $\alpha = \pi/2$} 
      \label{figure: 1_54_sequence_overlay}
   \end{figure}

Fig. \ref{figure: 1_54_sequence_overlay} shows the snapshots of the MorphoCopter traversing this shape 8 trajectories in the fully folded ultra-narrow configuration of $\alpha = \pi/2$. At this joint angle, the distance between the propellers and the roll axis is 0. Hence, there is no standard distance-based leverage available for the roll control. This verifies that the propellers' reaction moment is able to control the roll using the fixed inward tilt of the propellers and our adaptive controller design.

\subsection{Performance in a diamond shape trajectory in Experiments}
\begin{figure}[thpb]
      \centering
      
      \includegraphics[width=0.4\textwidth]{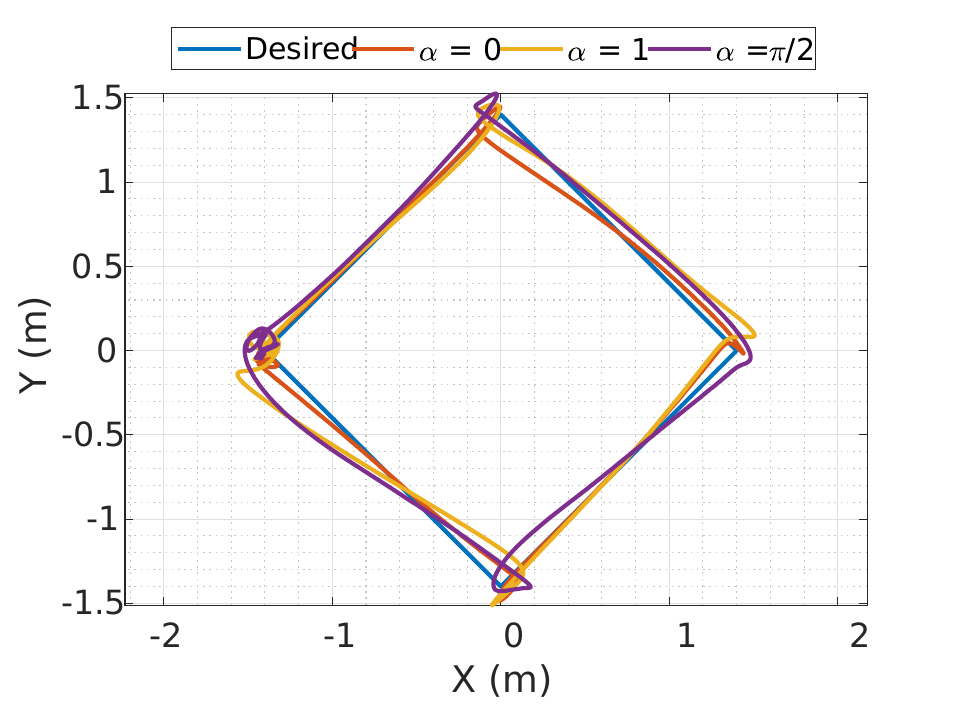}
      \caption{Diamond trajectory at various joint angles} 
      \label{figure: diamond_trajectory}
   \end{figure}

To further verify the controllability of the MorphoCopter in complex maneuvers, the experiments with a diamond-shaped trajectory were performed at various joint angles. The trajectory was a constant altitude (i.e., $-1.2\ m$) with X-Y trajectory in a diamond shape as shown in Fig. \ref{figure: diamond_trajectory}. The experiments were conducted with joint angles $\alpha =0, \alpha = 1$, and $\alpha = \pi/2$. This trajectory involves some sharp corners, which the MorphoCopter is able to traverse in a fully folded configuration as well. 

\begin{figure}[thpb]
      \centering
      
      \includegraphics[width=0.38\textwidth]{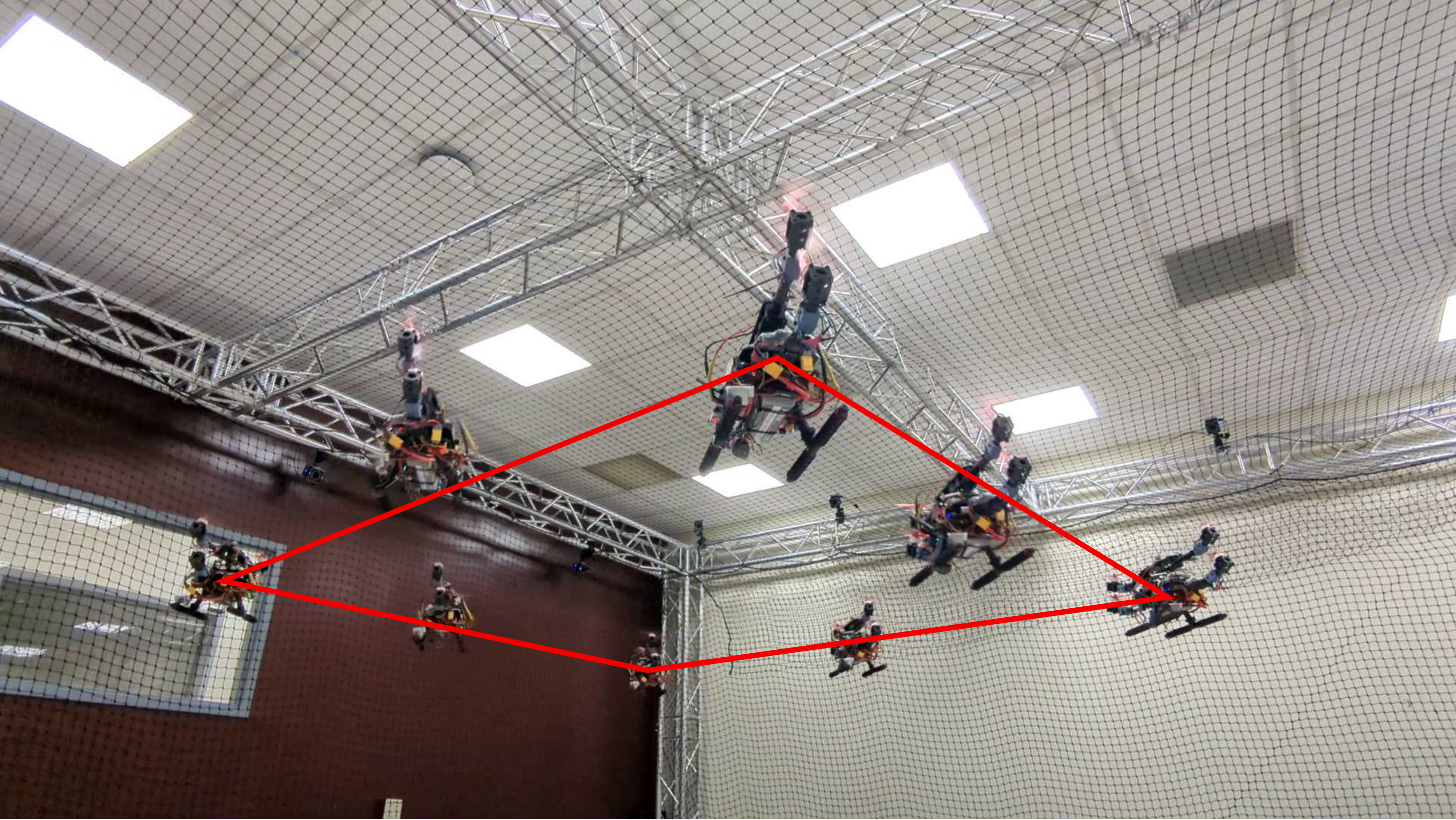}
      \caption{Snapshots from diamond trajectory traverse with $\alpha = \pi/2$} 
      \label{figure: diamond_sequence}
   \end{figure}

Fig. \ref{figure: diamond_sequence} shows the snapshots of the MorphoCopter traversing this diamond-shaped trajectory in a fully folded ultra-narrow configuration of $\alpha = \pi/2$. At this joint angle, again, it verifies the roll control using the propellers' reaction moment and our adaptive controller design.

\subsection{Passing through an environment with very narrow gaps}

\subsubsection{Simulation}

\begin{figure}[thpb]
      \centering
      
      \includegraphics[width=0.38\textwidth]{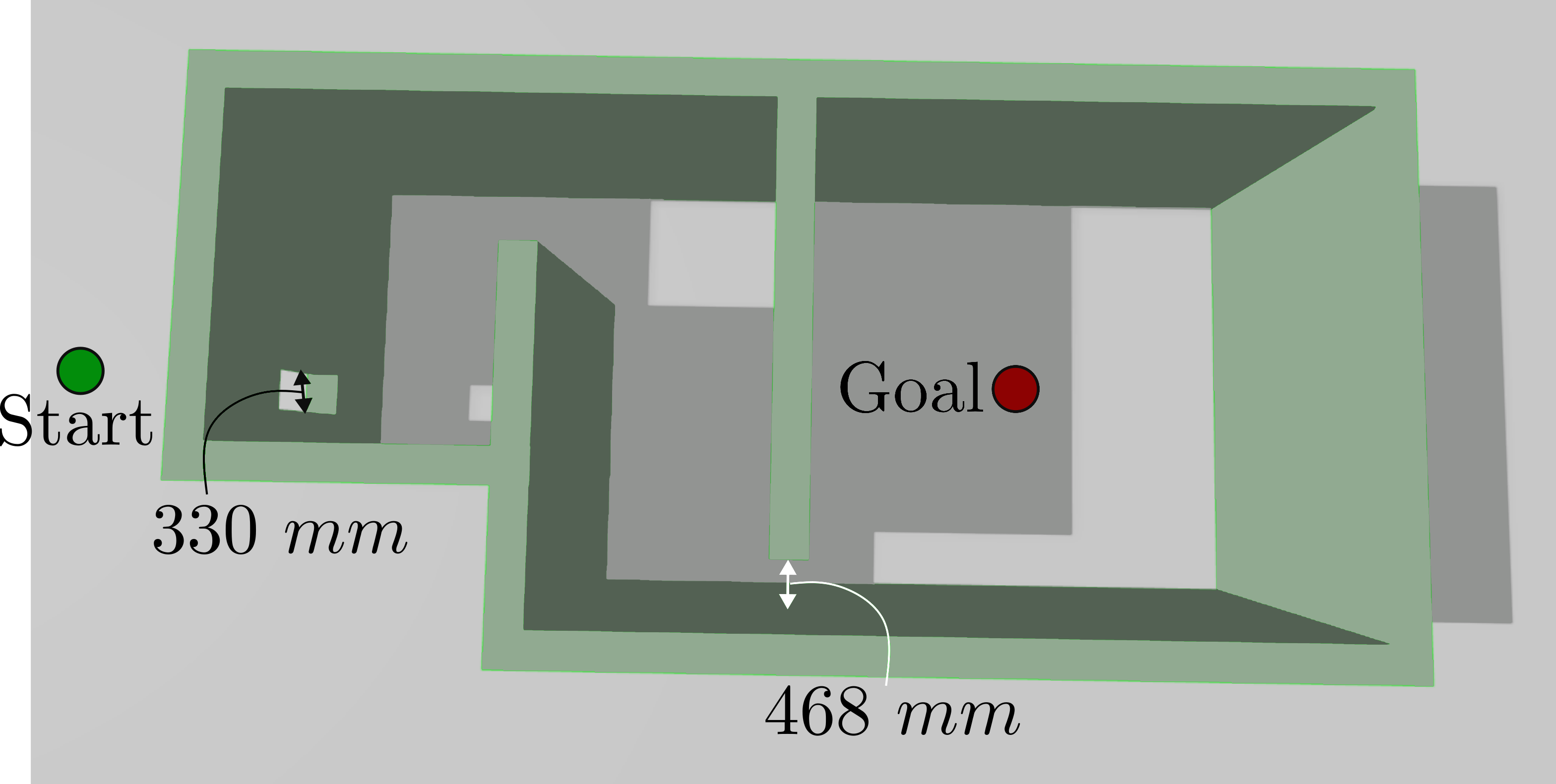}
      \caption{Simulation Environment} 
      \label{figure: environment}
   \end{figure}

\begin{figure}[thpb]
      \centering
      
      \includegraphics[width=0.38\textwidth]{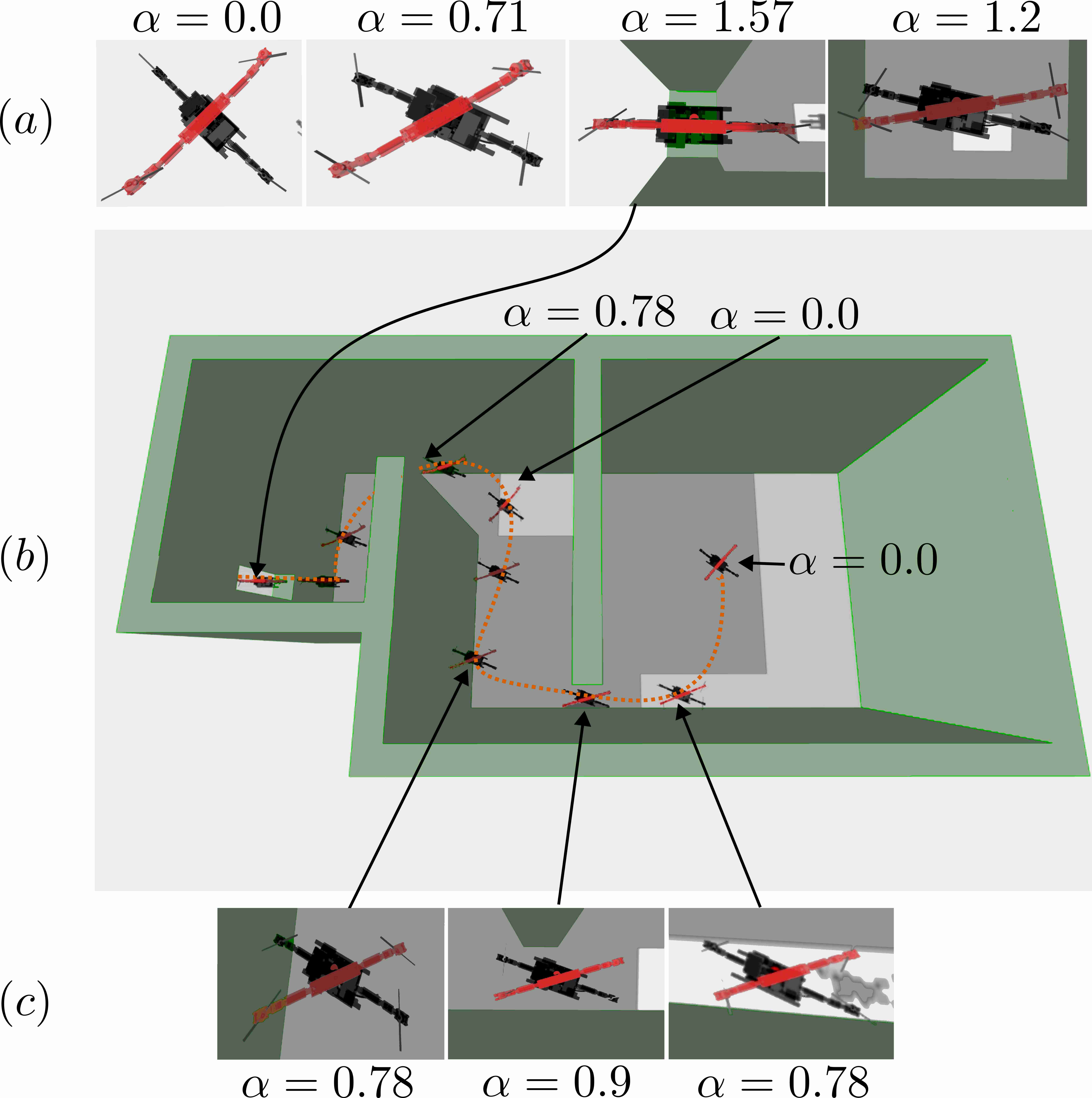}
      \caption{(a) Morphing sequence for passing through the first narrow gap of $330\ mm$ (b) Snapshots of the trajectory followed by MorphoCopter in simulation with multiple joint angle changes (c) Morphing sequence for passing through the second narrow gap of $468\ mm$} 
      \label{figure: simulation_trajectory_overlay}
   \end{figure}
   
First, we show the results from the simulation in a complex scenario containing multiple narrow gaps. Fig. \ref{figure: environment} shows the environment created for evaluating the performance of the MorphoCopter. The environment contains one narrow entrance with a width of $330\ mm$ and another narrow space with a width of $468\ mm$. The environment also contains walls that require the MorphoCopter to take multiple turns. As shown in Fig. \ref{figure: uav_angles}, the width of the MorphoCopter in a standard quadcopter configuration is $447\ mm$, which is larger than one of the narrow gaps and has a very small margin for the second narrow gap. Hence, even with a very precise controller, the standard quadcopter cannot access such an environment.

\begin{figure}[thpb]
      \centering
      \includegraphics[width=0.4\textwidth]{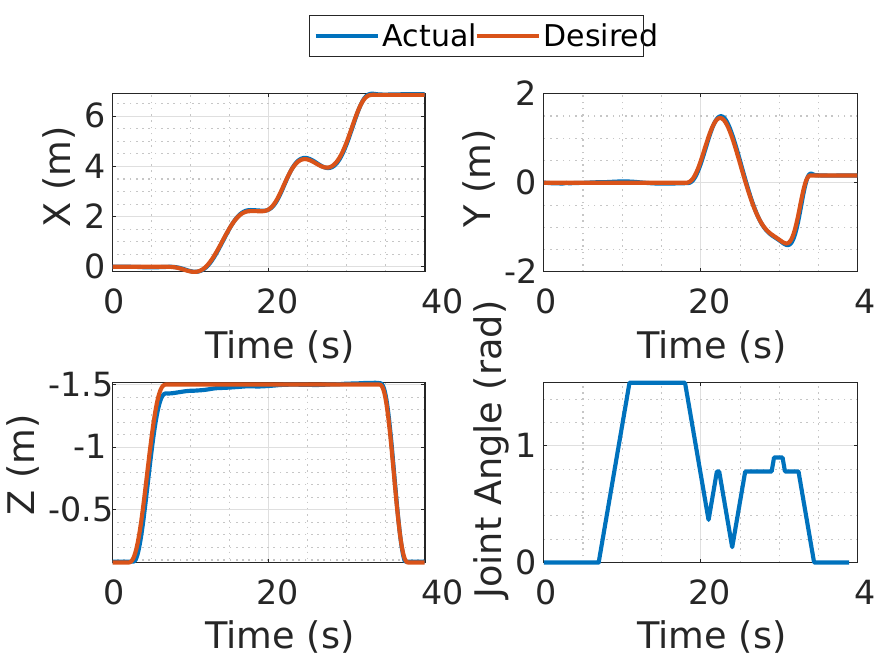} 
      \caption{Trajectory the MorhpoCopter followed while traversing the simulation environment} 
      \label{figure: simulation_plots} 
   \end{figure}

We design a minimum snap trajectory to pass through this environment, which involves multiple joint angle manipulations. Currently, the waypoints for the trajectory are determined manually. Fig. \ref{figure: simulation_trajectory_overlay} shows the trajectory followed by the MorphoCopter to pass through this environment along with a close up look at the morphing sequences for passing through narrow gaps. The dotted line refers to the trajectory in 3D space. Fig. \ref{figure: simulation_plots} shows the desired and actual trajectories in the time domain. We see that the MorphoCopter can change its configuration to various joint angles on the fly while passing through the environment. We are able to leverage the unique feature of the design by morphing the MorphoCopter only up to the desired angle without losing controllability.  

\subsubsection{Experiments}

\begin{figure}[thpb]
      \centering
      \includegraphics[width=0.48\textwidth]{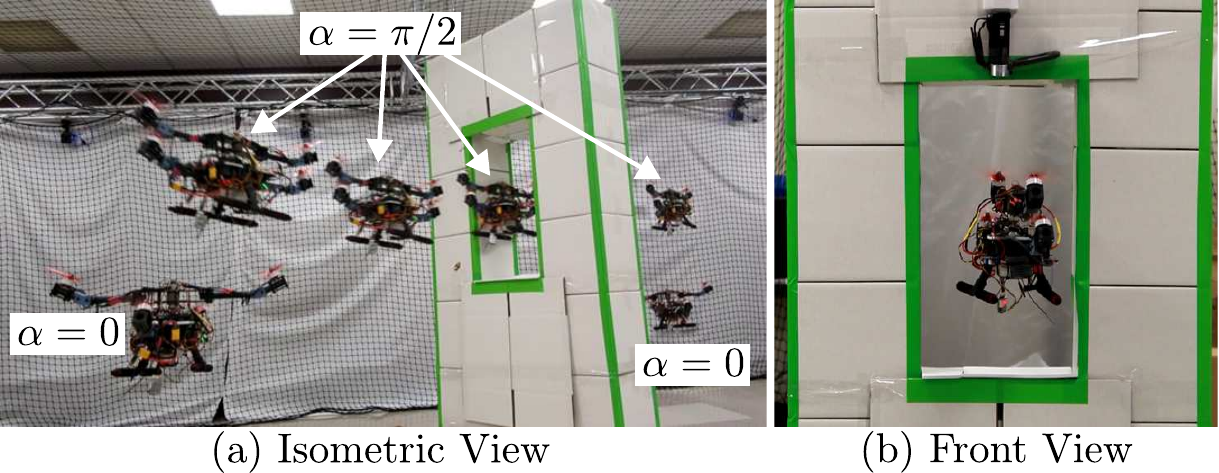}
      \caption{Snapshots from the video of the MorphoCopter passing through a narrow gap (a) Isometric view with snapshots taken at certain time intervals (b) Front view while the MorphoCopter is inside the narrow passage} 
      \label{figure: gap_passing_sequence}
   \end{figure}

As shown in Fig. \ref{figure: gap_passing_sequence}, a narrow passage was created using some obstacles. Due to space constraints, we test the performance of the MorphoCopter for the most difficult passage (i.e. $330\ mm$ wide) of the environment. 

\begin{figure}[thpb]
      \centering
      \includegraphics[width=0.45\textwidth]{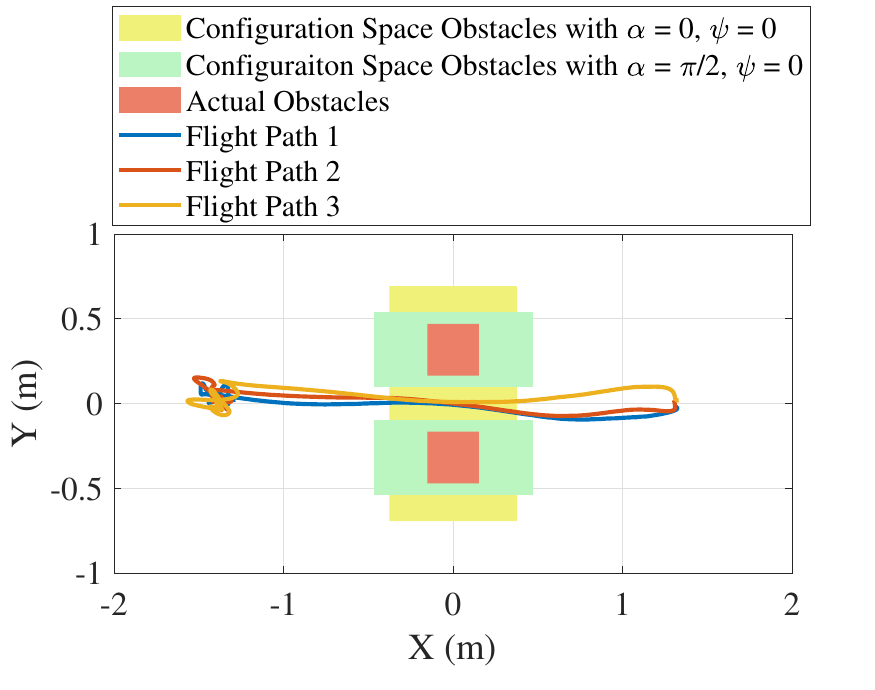}
      \caption{trajectory followed by the MorphoCopter to pass through the narrow gap at $\alpha {=} \pi/2$ along with configuration space obstacle representation at various joint angles} 
      \label{figure: gap_passing_path_xy}
   \end{figure}

\begin{figure}[thpb]
      \centering
      
      \includegraphics[width=0.31\textwidth]{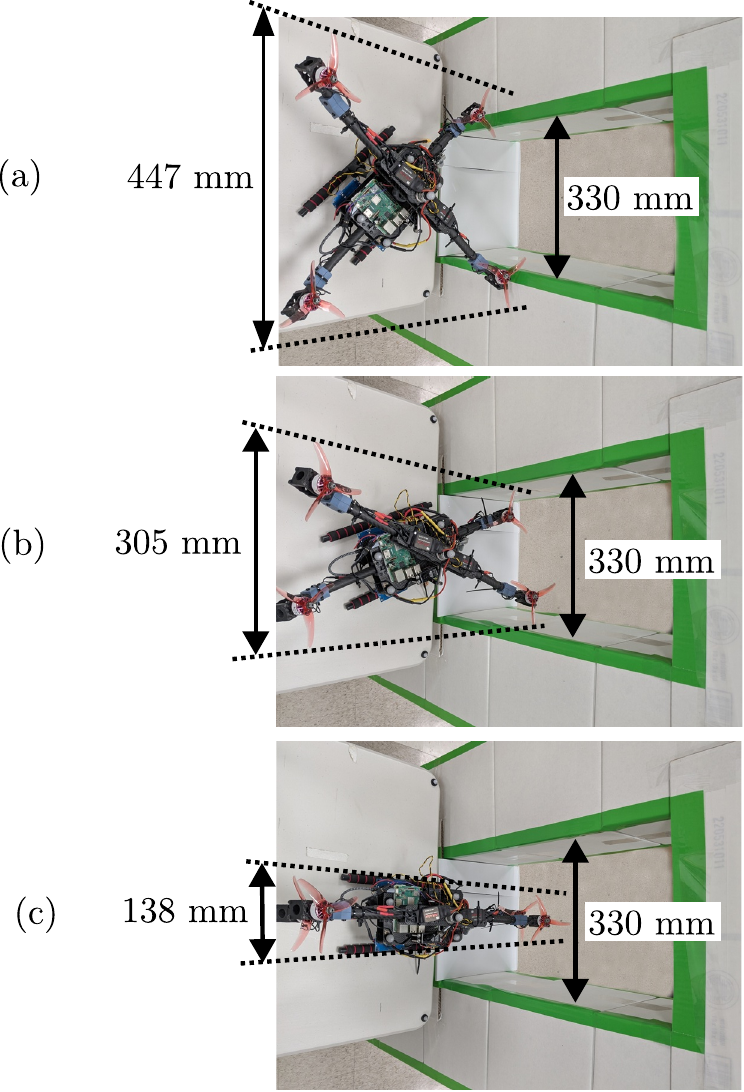}
      \caption{Visualization of MorphoCopter size at various joint angles against a narrow passage (a) $\alpha = 0$ (b) $\alpha = \pi/4$ (c) $\alpha = \pi/2$} 
      \label{figure: gap_passing_scale}
   \end{figure}

The difficulty of the standard quadcopter in passing through this narrow gap can be better visualized in the configuration space plot shown in Fig. \ref{figure: gap_passing_path_xy} and 3D space figures in Fig. \ref{figure: gap_passing_scale}. The plot shows the obstacles in an X-Y 2D plane with a red color. Considering the width of the MorphoCopter in standard X configuration ($\alpha = 0$), the configuration space obstacles are shown in yellow. As both obstacles are merged in the configuration space with $\alpha = 0$, the MorphoCopter cannot pass through them in a standard quadcopter configuration. The green boxes show the configuration space obstacle with the joint angle of $\pi/2\ rad$. Folding the MorphoCopter to this joint angle creates a passage of $166\ mm$ width for the MorphoCopter to pass through safely, leaving some room for errors in trajectory tracking. The actual path followed by the MorphoCopter in 3 different experiments is also shown in Fig. \ref{figure: gap_passing_path_xy}. 

\begin{figure}[thpb]
      \centering
      
      \includegraphics[width=0.43\textwidth]{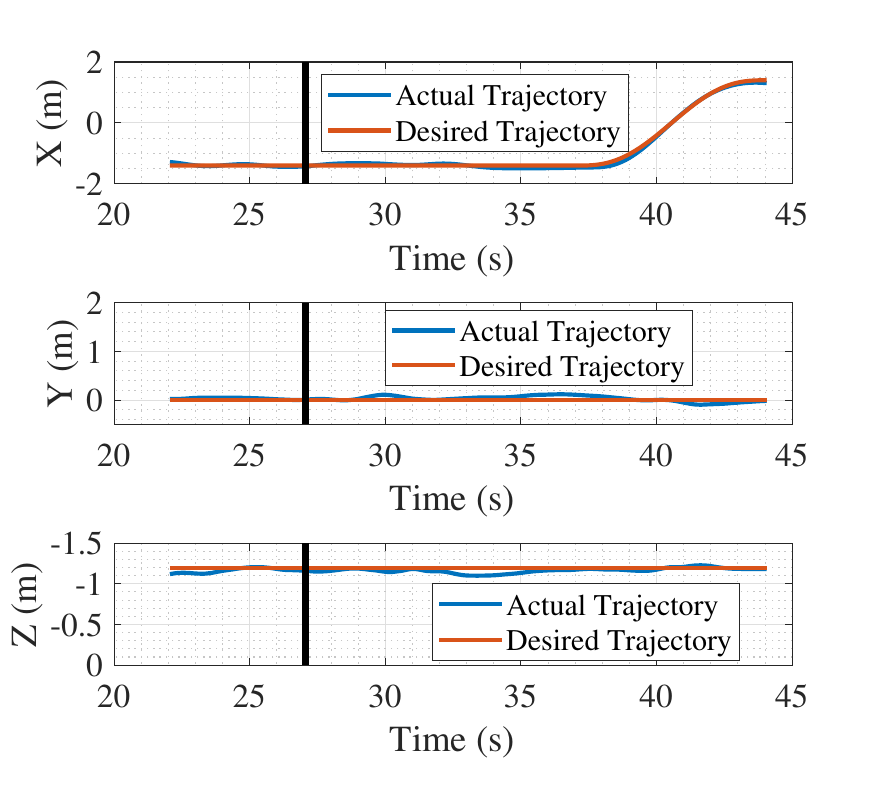}
      \caption{Trajectory of the MorphoCopter followed while passing through a narrow gap in an experiment. The vertical black line indicates the time when the MorphoCopter was folded} 
      \label{figure: gap_passing_trajectories}
   \end{figure}

Fig. \ref{figure: gap_passing_trajectories} shows the X, Y, and Z direction desired and actual trajectories for the MorphoCopter in the gap passing experiment. The black vertical line indicates the time when the MorphoCopter was commanded to fold from a standard quadcopter configuration to the fully folded bicopter configuration. Even with the narrow configuration after folding, it is able to track trajectories very well. 

\section{Conclusions}
\label{section:conclusion}
In this paper, we presented hardware design, adaptive controller, and experimental verification of the novel folding drone - MorphoCopter(4-2). We showcase that our simplistic design can reduce the size by the maximum amount of any available similar designs while still being controllable in all configurations. We verified the controllability of the MorphoCopter in any configuration through rigorous experiments and also showcased the ability to morph into a very narrow configuration, pass through a very narrow passage, and then come back to the standard configuration on the fly. Our current specific prototype has a limited flying duration and payload capacity, which we would like to improve in further iterations. Also, we will delve into designing optimum path planning and trajectory planning algorithms to leverage the capabilities of this novel MorphoCopter design.

\appendix[Hardware Details]
\label{appendix: hardware_details}
\begin{table}[htbp]
\caption{MorphoCopter(4-2) current design hardware details \label{table:parameters}}
\begingroup
\centering
\begin{tabular}
{|m{4.0cm}|m{3.5cm}|}
\hline 
\rowcolor{gray!20}mass ($m$) & $1.8\ kg$ \\
Propeller to C.O.M. distance ($l$) & $218.21\ mm$ \\ 
\rowcolor{gray!20} \relax [P, I, D] gains for x/y & [1.4, 1.4, 0.2]\\ 
$I_{ux}$ & $ 1.80 \times 10^{-3}\ kg\ m^2$\\
\rowcolor{gray!20} $I_{uy}$ &  $ 11.72 \times 10^{-3}\ kg\ m^2$\\
$I_{lx}$ &  $ 2.42 \times 10^{-3}\ kg\ m^2$\\
\rowcolor{gray!20} $I_{ly}$ &  $ 14.29 \times 10^{-3}\ kg\ m^2$\\
$K_m$ & $0.055\ Nm/N$\\
\rowcolor{gray!20}\relax [P, I, D] gains for z & [5.0, 4.5, 0.75]\\
\relax [P, I, D] gains for roll/pitch & [0.5, 0.075, 0.004]\\ 
\rowcolor{gray!20} \relax [P, I, D] gains for yaw & [0.2, 0.1, 0.004]\\ 
Motors & T Motor 2207 - 2500 KV \\
\rowcolor{gray!20}ESCs & T Motor Air - 40 A\\
Propellers & T Motor - 4943 tri-blade \\
\rowcolor{gray!20}Battery & HRB 3300 mAh - 4S LiPo 60C\\

\hline
\end{tabular}
~\newline
~\newline
C.O.M: Center of Mass
\endgroup
\end{table}

\FloatBarrier

\end{document}